\documentclass{egpubl}
\usepackage{VMV2023}
 
\WsPaper           

\usepackage[T1]{fontenc}
\usepackage{dfadobe}  

\biberVersion
\BibtexOrBiblatex
\usepackage[backend=biber,bibstyle=EG,citestyle=alphabetic,backref=true]{biblatex} 
\electronicVersion
\PrintedOrElectronic

\ifpdf \usepackage[pdftex]{graphicx} \pdfcompresslevel=9
\else \usepackage[dvips]{graphicx} \fi

\usepackage{egweblnk} 
\usepackage[T1]{fontenc}    
\usepackage{booktabs}       
\usepackage{amsfonts}       
\usepackage{amsmath}       
\usepackage{nicefrac}       
\usepackage{microtype}      
\usepackage{xcolor}         
\usepackage{subfiles}

\title{Exploring Physical Latent Spaces for High-Resolution Flow Restoration}

\author[C. Paliard et al.]
{\parbox{\textwidth}{\centering Chlo\'{e} Paliard$^{1}$, Nils Thuerey$^{2}$, Kiwon Um$^{1}$} 
        \\
{\parbox{\textwidth}{\centering $^1$LTCI, T\'{e}l\'{e}com Paris, Institut Polytechnique de Paris, France\\
         $^2$Technical University of Munich, Germany}
}
}

\begin{document}

\teaser{
    \centering
    \includegraphics[width=\linewidth]{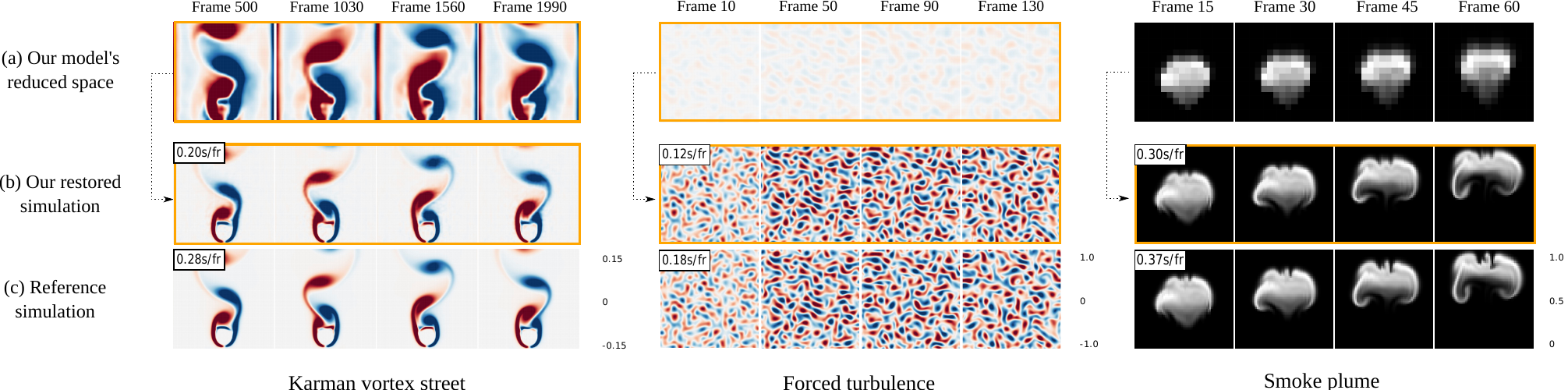}
    \caption{We propose a model composed of neural components and a physics solver, that autonomously discovers the reduced representation (a) that best fulfills the goal of restoring a fine reference simulation (b, c) from a unique coarse frame. This leads to relative improvements with respect to the baseline of 91\% on average for the Karman vortex street case, 74\% for the forced turbulence case and 35\% for the smoke plume case.}
    \label{fig:teaser}
}

\maketitle

\begin{abstract}
We explore training deep neural network models in conjunction with physics simulations via partial differential equations (PDEs), using the simulated degrees of freedom as latent space for a neural network. In contrast to previous work, this paper treats the degrees of freedom of the simulated space purely as tools to be used by the neural network. We demonstrate this concept for learning reduced representations, as it is extremely challenging to faithfully preserve correct solutions over long time-spans with traditional reduced representations, particularly for solutions with large amounts of small scale features. This work focuses on the use of such physical, reduced latent space for the restoration of fine simulations, by training models that can modify the content of the reduced physical states as much as needed to best satisfy the learning objective. This autonomy allows the neural networks to discover alternate dynamics that significantly improve the performance in the given tasks. We demonstrate this concept for various fluid flows ranging from different turbulence scenarios to rising smoke plumes.


\end{abstract}  


\section{Introduction}

Realistic simulation of natural phenomena is one of the ultimate goals of Computer Graphics research. Modeling and recreating such phenomena typically involves partial differential equations (PDEs) and thus numerical methods that aim for efficient computation of their solution. Despite the recent advances in numerical methods and computing power, many PDE problems of real-world scenarios, such as fluid simulation, require extremely costly calculations to resolve fine details, which are yet essential in the graphics context. Thus, using traditional numerical methods, which often require super-linear scaling in computation, remains challenging in practice. To minimize the costs of the computation, one can consider resolving the PDE in a reduced space, yet sacrificing the desired fine degrees of freedom both temporally and spatially. As an attractive compromise between computation and resolution, data-driven methods are becoming popular in many simulation problems \cite{morton2018deep,wiewel2020latent,kochkov2021machine}.

In this paper, we present a data-driven simulation technique for PDE problems with a novel training method that explores using physical states as latent space for deep learning. In contrast to many previous studies \cite{morton2018deep,kim2019,mohan2019compressed,stachenfeld2022learned}, our latent space is not composed of the output or intermediate states of a neural network, but is rather made of the physical states of a PDE solver, such as velocity fields. We train a deep neural network (DNN) to exploit the content of a reduced PDE solver and shape it in a way that best satisfies the given learning objective, i.e., achieving solutions that are as accurate as possible at our target high resolution space. This \emph{shaping} of the physical latent space gives the neural network a chance to discover modified dynamics and allows our model to better restore accurate high resolution solutions from them.
Examples of the reduced and restored solutions are shown in Fig.~\ref{fig:teaser}.

Our training method consists of an \emph{encoder} model transforming a coarse physical state using the degrees of freedom of a learnable latent space, a \emph{physics solver} corresponding to a given PDE followed by an \emph{adjustment} DNN, both operating in the reduced space, and a \emph{decoder} turning the reduced state into the target high resolution space. To train our models with a physics solver, we adopt a differentiable simulator approach \cite{hu2019difftaichi,holl2020,thuerey2021physicsbased}. We let the encoder model learn the latent space representation without any other constraint than the restoration of the target solution. Therefore, an end-to-end training of this pipeline gives the encoder the complete autonomy to shape the reduced representation.

This paper demonstrates that the autonomy of our training method leads to a better performance than previous work, especially in terms of generalization. We apply our method to various complex, non-linear PDE problems, based on the Navier-Stokes equations, which are essential in the context of modeling fluid flows. For all the scenarios, our model produces more accurate high-resolution results in a longer temporal horizon than conventional and more tightly constrained models.

\section{Related Work}

\paragraph*{Learning a PDE} The study of machine learning (ML) techniques for PDEs is getting more and more popular  \cite{crutchfield1987equations,kevrekidis2003equation,brunton2016discovering}. A conventional direction when using ML for PDEs is to aim for the replacement of entire PDE solvers by neural network models that can efficiently approximate the solutions as accurately as possible \cite{lusch2018deep,kim2019,wang2020physicsinformed,bhattacharya2021model}. In this context, Fourier Neural Operator \cite{li2021fourier} and Neural Message Passing \cite{brandstetter2022message} models have been introduced for learning PDEs, aiming at a better representation of full solvers with neural network models.

Instead of the pure ML-driven approach to solve target PDEs, an alternative approach exists in the form of hybrid methods that combine ML with traditional numerical methods. Among many different PDEs, fluid problems have received great attention due to their complex nature.
For example, a learned model can replace the most expensive part of an iterative PDE solver for fluids \cite{tompson2017,xiao2018adaptive} or supplement inexpensive yet under-resolved simulations \cite{um2018,sirignano2020dpm}. A regression forests model was also proposed for fast Lagrangian liquid simulations \cite{ladicky2015}.
For smoke simulations in particular, efficient DNNs approaches synthesize high-resolution results from low-resolution versions
\cite{chu2017,xie2018,bai2020dynamic} and convert low frame rate results into high frame rate versions \cite{oh2021two}.

\paragraph*{Differentiable solvers} Recently, differentiable components for ML have been studied extensively, particularly when training neural network models in recurrent setups for spatio-temporal problems \cite{amosKolter2017,NEURIPS2018_842424a1,toussaint2018differentiable,chen2018neural,schenck2018spnets,liang2019differentiable,wang2020differentiable,um2020solverintheloop,kochkov2021machine,zhuang2021learned}.
Consequently, a variety of differentiable programming frameworks have been developed for different domains
\cite{schoenholz2019jax,hu2019difftaichi,innes2019differentiable,holl2020}. 
These differentiable frameworks allow neural networks to closely interact with PDE solvers, which provides the model with important feedback about the temporal evolution of the target problem from the recurrent evaluations.
Targeting similar problems for temporal evolution, we employ a differentiable framework in our training procedure.

\paragraph*{Latent space representations} Effectively utilizing latent spaces lies at the heart of many ML-based approaches for solving PDEs. A central role of the latent space is to embed important (often non-linear) information for the given training task into a set of reduced degrees of freedom. For example, with an autoencoder network architecture, the latent space can be used for discovering interpretable, low-dimensional dynamical models and their associated coordinates from high-dimensional data \cite{champion2019datadriven}. Moreover, thanks to their effectiveness in terms of embedding information and reducing the degrees of freedom, latent space solvers have been proposed for different problems such as advection-dominated systems \cite{maulik2021reducedorder} and fluid flows \cite{wiewel2020latent,fukami2021sparse}. While those studies typically focus on training equation-free evolution models, we focus on latent states that result from the interaction with a PDE solver.
Neural network models have also been studied for the integration of a dynamical system with an ordinary differential equation (ODE) solver in the latent space \cite{chen2018neural}. This approach targets general neural network approximations with a simple physical model in the form of an ODE, whereas we focus on learning tasks for complex non-linear PDE systems.

\paragraph*{Reduced solutions} The ability to learn underlying PDEs has allowed neural networks to improve reduced, approximate solutions. Residual correction models are trained to address numerical errors of PDE solvers \cite{um2020solverintheloop}. Details at sub-grid scales are improved via learning discretizations of PDEs \cite{barsinai2019data} and learning solvers \cite{kochkov2021machine,stachenfeld2022learned} from
high-resolution solutions. Moreover, multi-scale models with downsampled skip-connections have been used for super-resolution tasks of turbulent flows \cite{fukami2019super}. These methods, however, typically employ a constrained solution manifold for the reduced representation. Indeed, the reduced solutions are produced using coarse-grained simulations with standard numerical methods, while our work shows the advantages of autonomously exploring the latent space representation through our joint training methodology.

\begin{figure*}[ht]
    \centering
    \includegraphics[width=\linewidth]{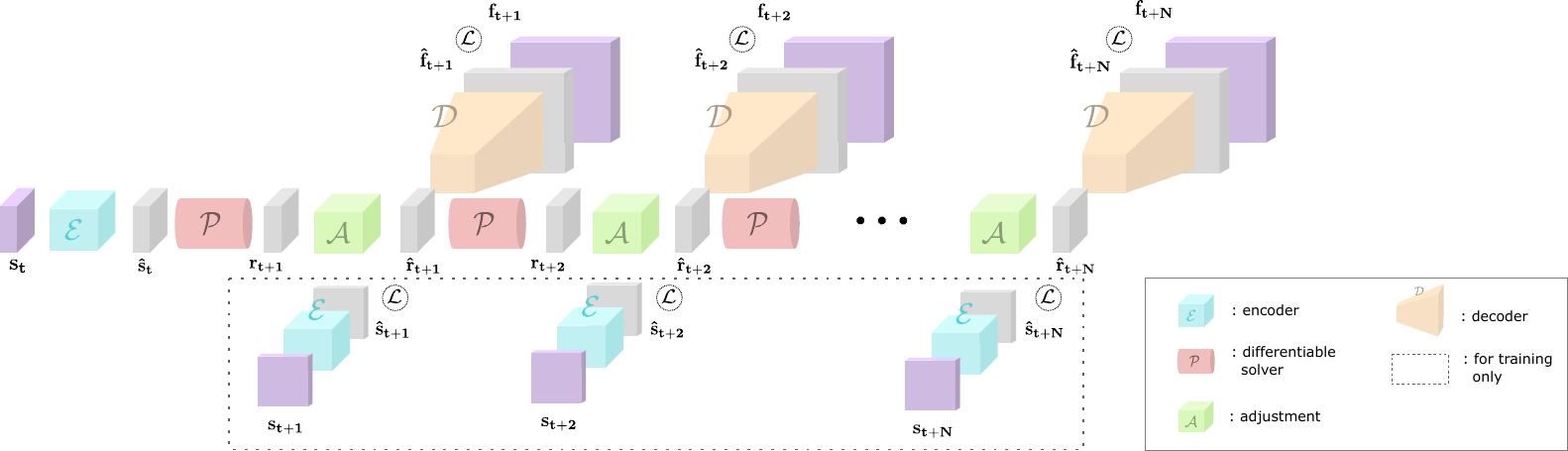}
    \caption{Architecture of our autonomous training approach 
    for $N$ integrated solver steps. The initial state is encoded into the latent space, the solver and adjustment models are applied $N$ times, and the adjusted states are decoded into the fine space.}
    \label{fig:architecture}
\end{figure*}

\section{Exploring Physical Latent Spaces}
\label{sec:models}

For a given learning objective, our training method explores how neural network models can leverage the physical states of a PDE as latent space.
Let $\mathbf{f} \in \mathbb{R}^{d_f}$ and $\mathbf{r} \in \mathbb{R}^{d_r}$ denote
two discretized solutions of a PDE, a fine and a coarse version respectively, with
$d_r \! \ll \! d_f$.
We focus on the numerical integration of this target PDE problem and indicate the
temporal evolution of each state as a subscript. A reference solution trajectory
integrated from a given initial state $\mathbf{f}_t$ at time $t$ for $n$ steps is represented by the finite set of states
$\{ \mathbf{f}_t, \mathbf{f}_{t+1}, \cdots, \mathbf{f}_{t+n} \}$. Each reference state is integrated over time with a fixed time-step size using a numerical solver $\mathcal{P}_f$, i.e.,
$\mathbf{f}_{t+1} = \mathcal{P}_f(\mathbf{f}_{t})$. Similarly, we integrate a
reduced state $\mathbf{r}_{t}$ over time using a corresponding numerical solver
$\mathcal{P}_r$ at the reduced space, which we will call \emph{reduced
  solver} henceforth,
i.e., $\mathbf{r}_{t+1} = \mathcal{P}_r(\mathbf{r}_{t})$. In this paper, we focus on cases where the solver $\mathcal{P}$ is the same for both reduced and fine discretizations.

Our model takes the linear down-sampling of $\mathbf{f}_t$, i.e., $\mathbf{s}_t = \emph{lerp}(\mathbf{f}_t)$, as input, and transforms it with the help of an encoder function $\mathcal{E}(\mathbf{s}| \theta_E): \mathbb{R}^{d_r} \rightarrow \mathbb{R}^{d_r}$, thus $\mathcal{E}(\mathbf{s}_t| \theta_E) = \mathbf{\hat{s}}_t$. Then, we can obtain the next reduced state $\mathbf{r}_{t+1} = \mathcal{P}(\mathbf{\hat{s}}_t)$.
Moreover, in order to keep the reduced solution consistent with the encoded representation over time, the output of the reduced solver is transformed by an adjustment function, $\mathcal{A}(\mathbf{r}_{t+1}| \theta_A) = \mathbf{\hat{r}}_{t+1}$. %
Thus, each reduced state $\mathbf{\hat{r}}_{t+i}$ is obtained by $i$ recurrent evaluations of the reduced solver and the adjustment function. 
Finally, a decoder function $\mathcal{D}(\mathbf{r}| \theta_D): \mathbb{R}^{d_r} \rightarrow \mathbb{R}^{d_f}$ 
restores a fine solution trajectory $\{ \mathbf{\hat{f}}_{t}, \mathbf{\hat{f}}_{t+1}, \cdots, \mathbf{\hat{f}}_{t+n}
\}$ from the reduced trajectory
$\{ \mathbf{\hat{r}}_{t}, \mathbf{\hat{r}}_{t+1}, \cdots, \mathbf{\hat{r}}_{t+n} \}$, thus $\mathbf{\hat{f}}_{t+i} = \mathcal{D}(\mathbf{\hat{r}}_{t+i}| \theta_D)$.
We model the  encoder, adjustment, and decoder functions as DNNs in which trainable weights are denoted by $\theta_E$, $\theta_A$, and $\theta_D$, respectively.

The joint learning objective of the three DNNs is to minimize the error between the approximate solutions and their corresponding reference solutions, i.e., $||\mathbf{\hat{f}}_{t+i} - \mathbf{f}_{t+i}||_2$. To guide the adjustment model, we additionally minimize $\|\mathbf{\hat{r}}_{t+i} - \mathcal{E}(\mathbf{s}_{t+i}| \theta_E)\|_2$. Thus, the final loss of our model is as follows:
\begin{equation}
\label{eq:loss}
\mathcal{L} = \sum_{i=1}^N \lambda_{hires} \times ||\mathbf{\hat{f}}_{t+i} - \mathbf{f}_{t+i}||_2 + \lambda_{latent} \times ||\mathbf{\hat{r}}_{t+i} - \mathcal{E}(\mathbf{s}_{t+i}| \theta_E)||_2 
\end{equation}
where $N$ denotes the number of integrated time-steps for training. Hence, at each training iteration, the gradients through all $N$ steps are computed for back-propagation and, consequently, all the models get jointly updated.

Fig.~\ref{fig:architecture} shows the architecture of our approach.
As the encoder does not receive any explicit constraint and has the complete freedom to \emph{autonomously} explore the reduced space to arrive at a suitable representation, we denote this approach by \emph{ATO}.

\paragraph*{Comparisons and baselines}

To illustrate the capabilities of our physical latent space, we compare \emph{ATO} to two state-of-the-art models
that operate in coarse space \cite{stachenfeld2022learned, um2020solverintheloop}. The
former, denoted by \emph{Dil-ResNet} in the following, represents a neural network model that 
aims at directly predicting solution states in the reduced space at each time-step. Hence, it does not make use of the reduced physics solver $\mathcal P$. On the other hand, the second variant \cite{um2020solverintheloop}, denoted by \emph{SOL}, consists of a differentiable physics solver and a trainable corrector model that addresses numerical errors of the solution states.
In both cases, unlike \emph{ATO}, the models are trained to make reduced solutions by targeting the linear down-sampling of the reference.

We note that these state-of-the-art models output solutions that stay in the coarse space. As our \emph{ATO} model aims at restoring high-resolution solutions using a decoder, a super-resolution model can transform these models' reduced solutions into high-resolution ones. To this end, we feed the reduced states produced by the \emph{Dil-ResNet} and \emph{SOL} models to a super-resolution network specialized for spatio-temporal turbulence problems \cite{fukami2019super}, resulting in high-resolution states. Henceforth, these models will be denoted as \emph{Dil-ResNet + SR} and \emph{SOL + SR}.

\section{Experiments}
\label{sec:experiments}

For each of the following scenarios, which are represented in 2D, the reference solution trajectories are generated for 200 steps from different initial conditions with a fixed time-step size. We focus on the velocity field for our restoration task and consider a four times coarser discretization for the reduced representation. More details about the experimental setups are given in the appendix.

\subsection{Karman vortex street}
\label{sec:karman}

We first consider a complex constrained PDE problem in the form of the Navier-Stokes equations. This problem is modeled as follows:
  \begin{align*}
    \label{eq:ns}
    \partial{\mathbf{v}}/\partial{t} = - (\mathbf{v} \cdot \nabla)\mathbf{v} -
    \nabla{p}/\rho + \nu \nabla^2\mathbf{v} \\
    \textrm{subject to} \quad \nabla \cdot \mathbf{v} = 0
  \end{align*}
where $\mathbf{v}$ is the velocity, $p$ is the pressure, $\rho$ is the density and $\nu$ is the viscosity.

\begin{figure}[t]
    \centering
    \includegraphics[width=0.7\linewidth]{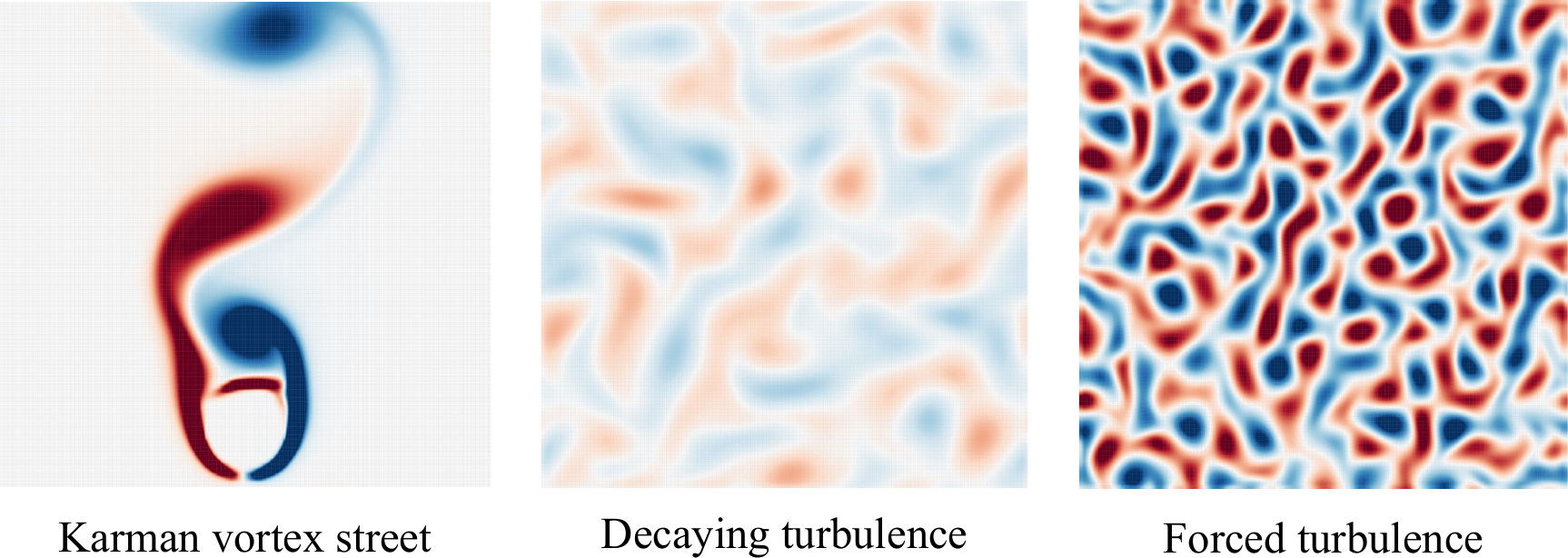}
    \caption{Example frames from the test data-set of each scenario.}
    \label{fig:test_set}
\end{figure}

In this scenario, shown in Fig.~\ref{fig:test_set} (left), a continuous inflow collides with a fixed circular obstacle. It creates an unsteady wake flow, which evolves differently depending on the Reynolds number. For the reference solutions, we use a numerical fluid solver that adopts the operator splitting scheme, Chorin projection for implicit pressure solve \cite{chorin1967numerical}, semi-Lagrangian advection \cite{stam1999}, and explicit integration for diffusion. We choose Reynolds numbers between $Re=90$ and $Re=1190$ for our training data-set, and Reynolds numbers from $Re=450$ to $Re=1400$ for testing. The encoder of our \emph{ATO} setup takes the Reynolds number as an additional input in order to guide the exploration of the reduced space. The \emph{Dil-ResNet} and \emph{SOL} setups also receive the Reynolds number to let the models learn different physics evolutions. Finally, in order to be fair in our comparisons, the obstacle mask is applied to each state output by the \emph{Dil-ResNet} solver.
The domain of all target solutions is discretized with $128\times256$ cells and has a staggered layout with closed boundaries, except for the bottom boundary for the constant inflow velocity and the top boundary that remains open. Although this setup targets a periodic evolution, the models are trained on more than one period in order to see a variety of initial states.

\subsection{Decaying and forced turbulence}
 \label{sec:turb}

This scenario is likewise based on the Navier-Stokes equations and is initialized with vortices randomly placed in the domain. We present two different sub-cases, one where the vortices are decaying over time without any external influence, and the other where an external force field $\mathbf{g}$ slows down the decaying motion of the vortex structures, enabling the creation of richer dynamics. The viscosity $\nu = 0.1$ is used in both training data-sets, and the multiplicity of initial velocities and force fields enables a great variety of evolutions. Example frames of the decaying turbulence case are shown in Fig.~\ref{fig:test_set} (middle). In the forced turbulence scenario, shown in Fig.~\ref{fig:test_set} (right), as an additional degree of freedom for our learnable latent representation, we let the encoder of the \emph{ATO} setup infer a latent force field $\mathbf{\hat{g}}$ that is integrated by the reduced solver. The linearly down-sampled field $\mathrm{lerp(\mathbf{g})}$ is used in the other setups.
Each solution's domain is discretized with $128^2$ cells and has a centered layout with periodic boundary conditions. A different randomized force sequence is generated for each simulation of the forced turbulence case, and the test sets are generated using different initial vortices and force fields than for training.

\subsection{Smoke plume}

This last scenario serves as a proof-of-concept of our method for more
complex, practical graphics applications. Aiming for complex flow behavior driven by hot smoke plumes, we set up an initial smoke volume as a marker field with an arbitrary density distribution in a circular shape. The marker field is then passively advected by the velocity field and, at the same time, induces a buoyancy force via the Boussinesq approximation, that is influencing the velocity evolution. Therefore, the marker and velocity fields are tightly coupled. This scenario considers a more challenging problem of the Navier-Stokes equations than before, naturally making the fluid flow more interesting and providing a harder task for our model.
The training and test data-sets are composed of smoke volumes initialized with random noise, with a fixed radius and position. The passive marker field is given as an input to our encoder and adjustment models, but its linearly down-sampled version is used in the reduced solver. Then, only the velocity field is up-sampled, and the high-resolution marker field is advected by the predicted velocity.
The simulation domain is discretized with $128^2$ cells adopting a centered layout for the marker field, a staggered layout for the velocity field, and open boundary conditions.

\subsection{Network architecture and training procedure}
\label{sec:archi}

The encoder $\mathcal{E}(\cdot | \theta_E)$ is implemented as a simple convolutional neural network (CNN) and the adjustment model is composed of convolutional layers that are interleaved with skip-connections. For the decoder, we adapt the multi-scale model for turbulent flows \cite{fukami2019super}, and the separate super-resolution network used in \emph{Dil-ResNet + SR} and \emph{SOL + SR} employs this same model. For all models, every convolutional layer except for the last one is followed by the Leaky ReLU activation function, except for \emph{Dil-ResNet} which uses ReLU activations. We adopt circular padding for the periodic boundary condition problems and zero padding for the others. The architectures of the models are detailed in the appendix.

At each training iteration, for a given batch size, we randomly sample the initial states from the reference solution trajectories and integrate the approximate solution trajectories for $N$ steps. All our trainings use an Adam optimizer \cite{kingma2014adam} and a decaying learning rate scheduling.

\section{Results}
\label{sec:results}

We evaluate the trained models based on relative improvements over a \emph{baseline} simulation. To make the baseline solutions, we apply the reduced solver without interactions with any neural network and up-sample the reduced states into the reference space with a linear interpolation. Errors are computed with respect to the reference solutions, hence an improvement of 100\% would mean that the restored solutions are identical to the reference. We evaluate each model using the mean absolute error (MAE) and mean squared error (MSE) metrics, which we measure in both velocity and vorticity. We present the results of the models trained with the highest number of integrated steps for each scenario, as they show better performance in general.

\subsection{Reduced representations}
\begin{figure}[t]
    \centering
    \includegraphics[width=\linewidth]{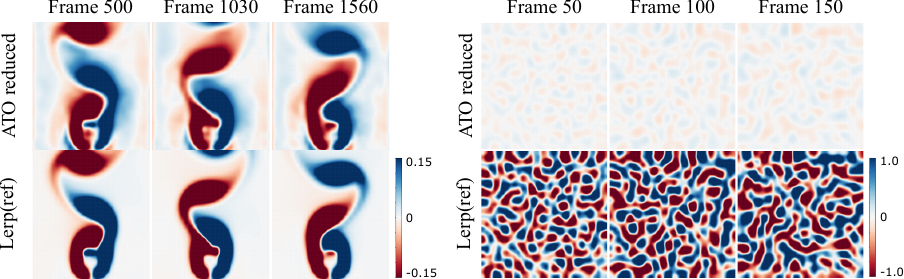}
  \caption{Reduced frames for the Karman vortex street case (left), and the forced turbulence case (right). The latent space of \emph{ATO} (top) and the linearly down-sampled reference (bottom) are shown.}
  \label{fig:latent_images}
\end{figure}

\begin{figure}[t]
    \centering
    \includegraphics[width=0.49\linewidth]{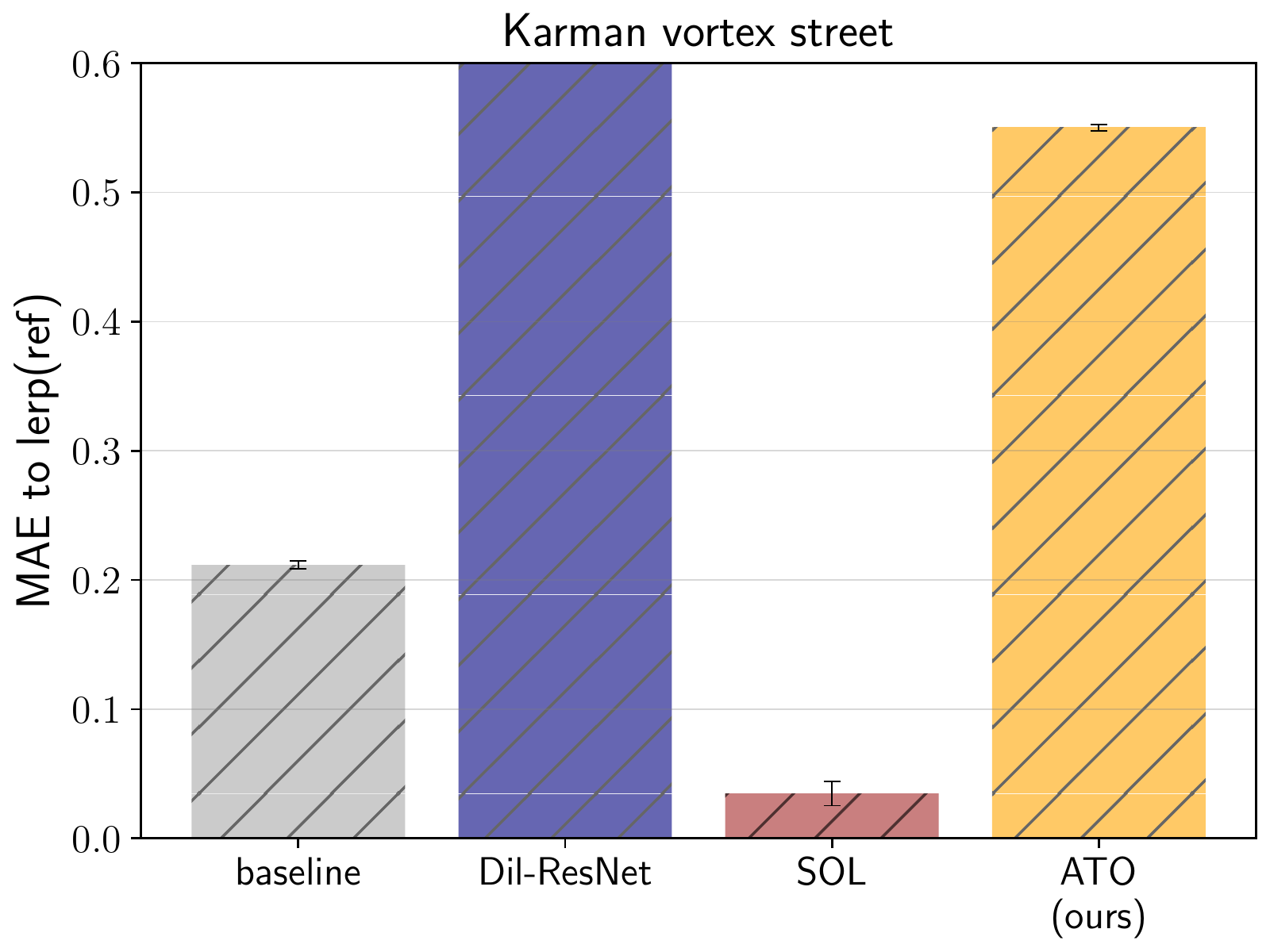}
    \includegraphics[width=0.49\linewidth]{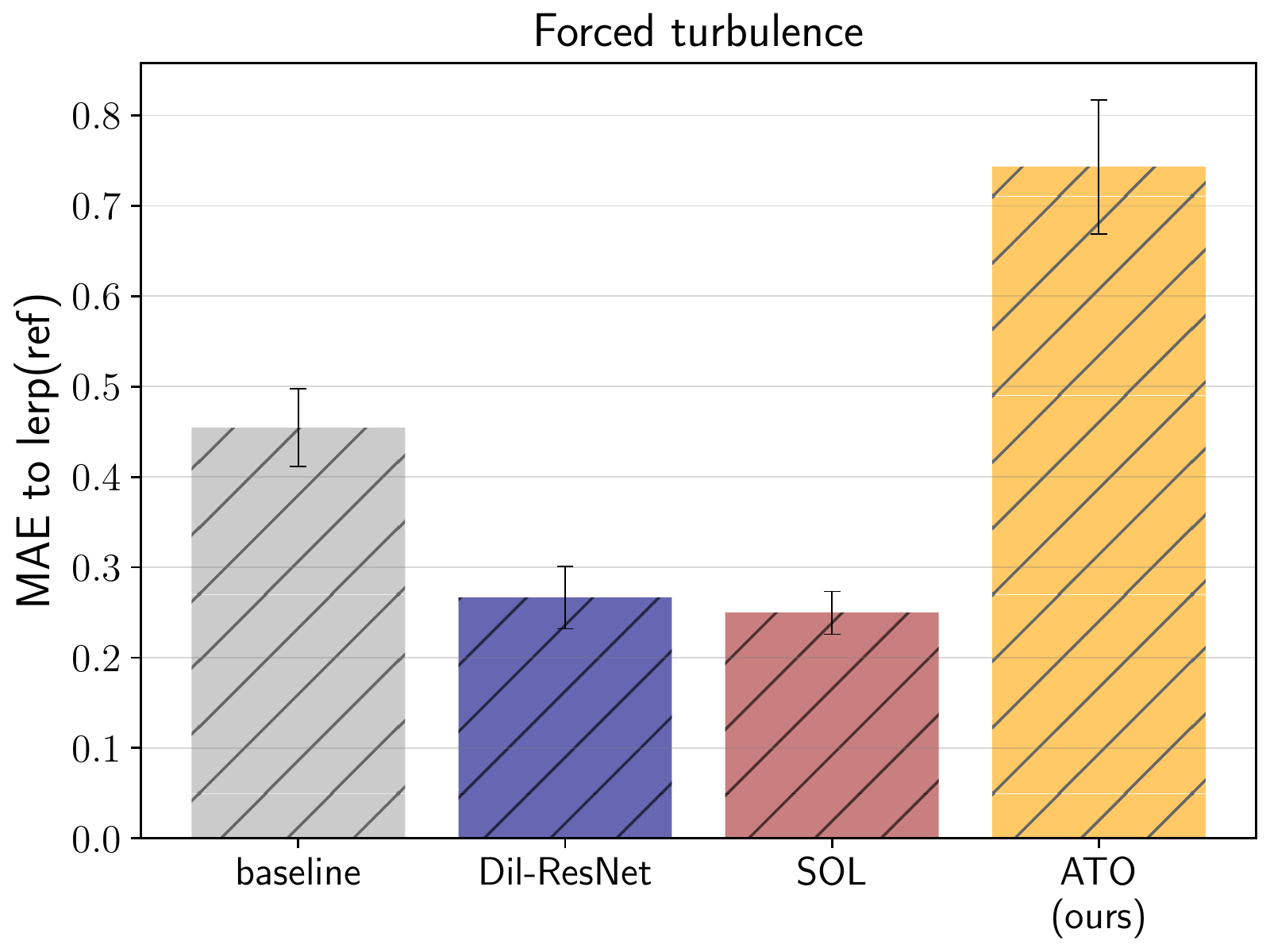}
    \caption{Distance between each model's reduced space and the down-sampled reference. The error bars indicate the standard deviation over the test runs.}
    \label{fig:latent_plot}
\end{figure}

The images of Fig.~\ref{fig:latent_images} show visual examples of the reduced representations for the Karman vortex street and forced turbulence scenarios, for different time-steps. The graphs of Fig.~\ref{fig:latent_plot} show the quantified differences between the reduced states produced by the different trained models and the conventionally down-sampled reference states. We observe that our training procedure leads the latent representation to have vortex structures that are very similar to conventional down-sampling, while being considerably different quantitatively. Thus, we believe that the reduced representation of the \emph{ATO} model stays physically meaningful for the numerical solver yet adds signals for accurately decoding high resolution states. We note that different training initializations of the same scenario produce latent representations that stay close to each other, which indicates that there exists a manifold of latent solutions that our \emph{ATO} model converges to in order to get the best performance.

\subsection{Karman vortex street}
\label{sec:results_karman}
This example considers different vortex shedding behaviors depending on the Reynolds number of each simulation.
We evaluate the models trained with 16 integrated steps on six test simulations with Reynolds numbers ranging from 450 to 1400, consisting of 2000 time-steps each.
In this scenario, we test the extrapolation ability of the models both physically and temporally, with higher Reynolds number thus more turbulent simulations than for training, and ten times longer sequences.

Table~\ref{perf_table} shows that \emph{ATO} outperforms the other models, with a relative improvement of 91\% (and 88\%) in terms of velocity MAE (and vorticity), while \emph{SOL + SR} improves the baseline by 84\% (and 83\%). On the other hand, the \emph{Dil-ResNet + SR} model fails to retrieve the target simulation for more than 200 time-steps, and thus is incapable of generalization in this scenario. The temporal metrics shown in the appendix demonstrate the capability of \emph{ATO} to correctly restore a solution for longer time ranges than the other models. More specifically, the distance between the reduced states and \emph{lerp(ref)} show that the \emph{ATO} model is the only one to have a consistent latent representation over time, which proves its better temporal extrapolation capabilities.

Fig.~\ref{fig:error_maps} (left) shows examples of high-resolution frames produced by each model along with the spatial distribution of the absolute error in velocity, for Re = 1400. Although the visual quality of the different results seems equivalent at first sight, one can notice that the position of the vortices is more accurate in \emph{ATO}'s outputs than \emph{SOL+SR}'s.
These results show that our \emph{ATO} model, thanks to its latent space content, has learned to approximate the physical dynamics of the simulation more accurately.

\begin{table}[hb]
  \centering
  \resizebox{\linewidth}{!}{
  \begin{tabular}{cccccccccc}                 \\
    \toprule
    & \multicolumn{3}{c}{Karman vortex street} & \multicolumn{3}{c}{Decaying turbulence} & \multicolumn{3}{c}{Forced turbulence} \\
    & \multicolumn{3}{c}{($128\times256$)} & \multicolumn{3}{c}{($128\times128$)} & \multicolumn{3}{c}{($128\times128$)} \\
    \cmidrule{2-10}
     & MAE & MSE & runtime & MAE & MSE & runtime & MAE & MSE & runtime \\
    \midrule
    Reference   & N/A & N/A  & 28.2   & N/A & N/A & 14.1  & N/A & N/A & 17.9 \\
    \midrule
    Baseline    & 0.214 & 0.096  & 11.1  & 0.069  & 0.024 & 7.7 & 0.504 & 0.426 & 9.6 \\
    \midrule
    Dil-ResNet+SR  & 3580 & 1407 & 9.2  & 0.033  & 0.023 & 3.8  & 0.272 & 0.167 & 5.3  \\
    \midrule
    SOL+SR     & 0.035 & 0.005 & 20.0  & 0.013 & 0.005 & 12.8  & 0.256 & 0.137 & 14.2 \\
    \midrule
    \textbf{ATO (ours)}  & \textbf{0.020} & \textbf{0.002} & 20.4  & \textbf{0.012} & \textbf{0.005} & 11.5  & \textbf{0.133} & \textbf{0.040} & 12.4  \\
    \bottomrule
    \\
  \end{tabular}
  }
  \caption{Summary of the MAE and MSE metrics, along with the runtime for one simulation of 100 frames (averaged over ten runs, in seconds).}
  \label{perf_table}
\end{table}

\subsection{Decaying turbulence}

In this example, we consider initially chaotic turbulent flows that slowly decay over time. We evaluate the models trained with 16 integrated steps, on five random initializations, lasting 200 steps each.

Table.~\ref{perf_table} shows that the \emph{ATO} model yields greatly improved results with a relative improvement of 83\% (and  80\%) in terms of velocity MAE (and vorticity). However, in this more simple case, the \emph{SOL + SR} model also improves the baseline significantly with 82\% (and 78\%) of relative improvement. \emph{Dil-ResNet + SR}, however, only yields 53\% (and 6\%) of improvement. Examples frames for this scenario are shown in the appendix.

\subsection{Forced turbulence}

This complex fluid flow scenario considers the same experimental setup as in the previous case but with external forces, which leads to highly chaotic turbulent flows. We evaluate the models trained with 16 integrated steps, on five random initializations both in velocity and forcing, for 200 steps.

\begin{figure*}[ht]
  \centering
  \includegraphics[width=\linewidth]{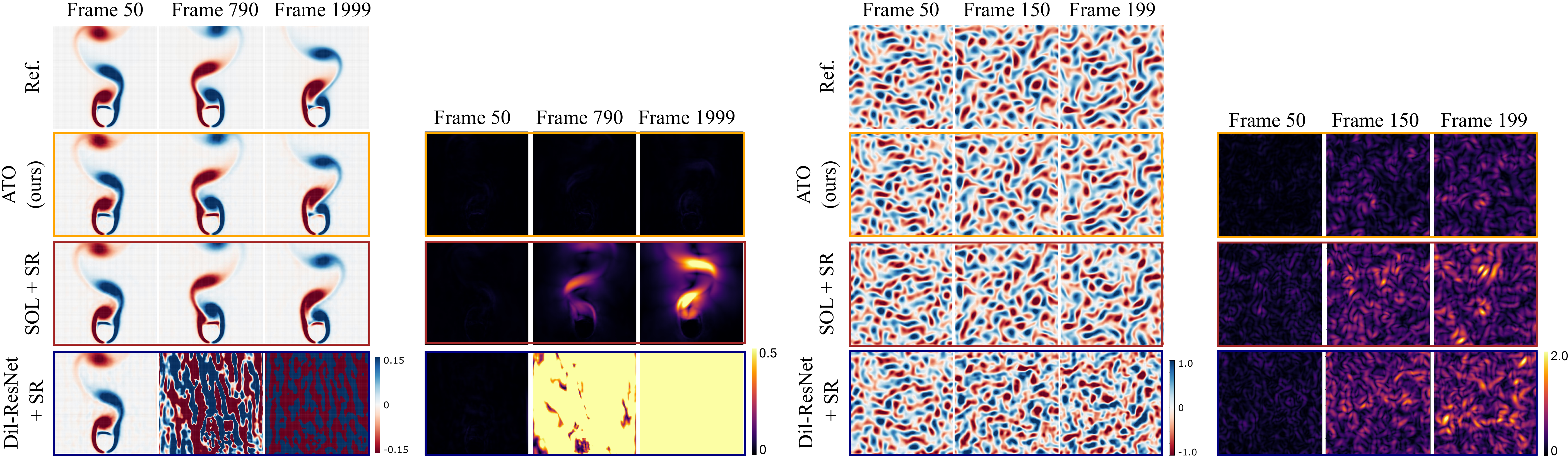}
  \caption{Example frames for all models along with the absolute error in velocity are shown for the Karman vortex street case with Re = 1400 (left) and for the forced turbulence case (right).}
  \label{fig:error_maps}
\end{figure*}

Table~\ref{perf_table} shows that the \emph{ATO} model significantly improves the baseline with a relative improvement of  74\% (and  69\%) in terms of velocity MAE (and vorticity). In comparison, \emph{SOL + SR} improves by only 49\% (and 43\%) and \emph{Dil-ResNet + SR} by 46\% (and 38\%). Therefore, in this complex case with external forcing and more turbulent flows, the \emph{ATO} model particularly stands out.
Fig.~\ref{fig:error_maps} (right) shows examples of high-resolution frames produced by each model along with the spatial distribution of the absolute error in velocity.

\subsection{Smoke plume}
 \label{sec:smoke_plume}

In this last example, we consider complex flow behaviors created by hot smoke plumes that evolve from random circular densities. We evaluate our model trained with 32 integrated steps on five test simulations with different initial marker fields from which we perform a "warm-up" of 50 time-steps, in order to get interesting plume shapes.

\begin{figure}[ht]
  \centering
  \includegraphics[width=\linewidth]{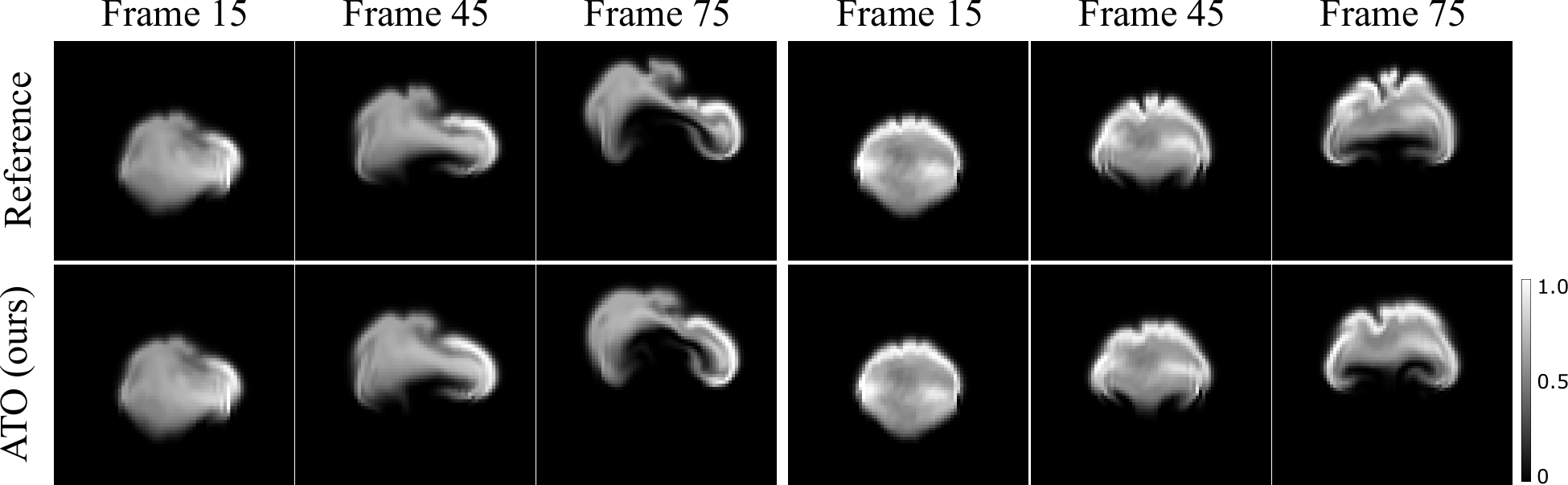}
    \caption{Example frames of the smoke plume scenario for two different initializations, for the \emph{ATO} model.}
  \label{fig:smoke_plume}
\end{figure}

Fig.~\ref{fig:smoke_plume} shows that, despite the increased difficulty of this challenging scenario, our \emph{ATO} model succeeds at reconstructing a complex high-resolution plume of good quality. Indeed, our method presents an improvement of 35\% on average over the baseline for 100 steps.

\subsection{Ablation study}

In order to see the effects of each of its components, we evaluate our \emph{ATO}
model with differently ablated training setups for the forced turbulence
scenario.
Our ablation study includes the following models:
\begin{itemize}
\item No latent loss: we remove the second term of the loss in
  Eq.~\ref{eq:loss}; consequently, our training does not constrain the adjusted
  states to match the encoder-induced latent space.
\item No encoder: we omit the encoder such that the latent
  representation is constrained to be conventional linear down-sampling.
\item No encoder \& no latent loss: since the previous model's reduced space is constrained to linear down-sampling, we test the same setup without the encoder and with no latent constraint.
\item No solver: we replace the \emph{solver + adjustment} part of our
  \emph{ATO} model with the \emph{Dil-ResNet} NN-solver in order to study the effect
  of a non-physical latent space.
\item No adjustment: we evaluate a setup where the reduced simulation evolves
  without being adjusted.
\item \emph{lerp(forces)}: we input a simple linear down-sampling of the force
  fields to the reduced solver, instead of their encoded representation.
\end{itemize}

\begin{table}[ht]
  \centering
  \resizebox{0.93\columnwidth}{!}{
  \begin{tabular}{cccccc}                
    \toprule
     & \multicolumn{2}{c}{Velocity} & \multicolumn{2}{c}{Vorticity} & Latent space \\
     \cmidrule{2-6}
     & MAE & MSE & MAE & MSE & MAE \emph{lerp(ref)} \\
    \midrule
    ATO (ours)              & 0.133 & 0.040  & 0.084 & 0.015  & 0.743 \\
    \midrule
    no latent loss          & 0.156 & 0.057  & 0.097 & 0.020  & 0.488 \\
    \midrule
    no encoder              & 0.259 & 0.146  & 0.156 & 0.48  & 0.253 \\
    \midrule
    no enc. \& no lat. loss & 0.284 & 0.174  & 0.167 & 0.055  & 0.322 \\
    \midrule
    no solver               & 0.737 & 1.073  & 0.578 & 0.651  & 0.713 \\
    \midrule
    no adjustment           & 0.409 & 0.339  & 0.212 & 0.085  & 0.568 \\
    \midrule
    \emph{lerp(forces)}     & 0.944 & 1.725  & 0.429 & 0.325  & 0.682 \\
    \bottomrule
    \\
  \end{tabular}
  }
  \caption{Results of the ablation study: we present the MAE and MSE in velocity and vorticity for each model, along with the distance between its reduced space and the down-sampled reference.}
  \label{ablation_table}
\end{table}

Table~\ref{ablation_table} shows that the encoder, physics solver, and adjustment components of our \emph{ATO} model are essential for its good performance. Firstly, the \emph{no encoder} and \emph{no encoder \& no latent loss}
experiments confirm that, with \emph{lerp(ref)} as initial reduced representation and without our encoder, the adjustment network was not able to find a latent representation that would lead to an optimal performance. Furthermore, the \emph{no latent loss} ablation shows that the latent loss guiding the adjustment model via the encoder results in a better performance.
Note that the performance of \emph{ATO} significantly decreased when the encoder was
absent, whereas the performance drop due to omitting the latent loss was relatively less significant.
Secondly, the \emph{no solver} and \emph{no adjustment} experiments show that
using a reduced physics solver in conjunction with an adjustment model is crucial
for the good performance of our \emph{ATO} model.
Finally, the \emph{lerp(forces)} experiment indicates that our encoder model failed to find a latent representation for the velocity that was compatible with an external factor conditioned to \emph{lerp}.

All of the models tested in this ablation study gave comparable
standard deviation values within the test set; thus, we did not include them in the table.

\subsection{Runtime performance}
For each scenario, we compare the runtime performance of the trained models with the reference's, measuring timing for one simulation of 100 frames, averaged over ten different runs. For \emph{ATO}, the computations start with the initial velocity inference by the encoder model and stop when all 100 frames are output by the decoder. All timings were computed using a single \emph{GeForce RTX 2080 Ti} with 11GiB of VRAM.

Table~\ref{perf_table} shows the summary of computational timings for the reference, baseline (reduced solver without any DNN model), and trained models. For all four cases, our \emph{ATO} model yields improvement in runtime compared to the reference. For the Karman vortex street scenario, our \emph{ATO} model speeds up the computations by 28\%, against 29\% for \emph{SOL+SR}. Yet, as shown in Sec~\ref{sec:results_karman}, \emph{ATO} shows an improvement of the baseline MAE that is 7\% better than \emph{SOL+SR}. Similarly, for the forced turbulence case, the \emph{ATO} model speeds up the computations by 18\%, against 9\% for \emph{SOL+SR}, and improves the baseline MAE of 25\% more than \emph{SOL+SR}. For the smoke plume scenario, our ATO model speeds up the reference by 20\% compared to 28\% for the baseline, while improving the baseline MAE by 35\%. We note that the \emph{Dil-ResNet + SR} model often has the best runtime performance because it does not contain any numerical solver, but it has errors at least 50\% higher than \emph{ATO} and shows very poor temporal extrapolation capabilities.

Training our \emph{ATO} model takes between one and three days depending on the physical scenario, on a \emph{Tesla V100} with 16GiB of VRAM.

\vspace{1cm}

\section{Limitations and Future Work}
\label{sec:limit}

These results show that our training method using the states of physics simulations as latent space of DNNs can facilitate the learning task for complex simulations. This provides a starting point for the exploration of physical latent spaces in many different problems. However, we note that our \emph{ATO} model is not particularly standing out in a simple scenario like the decaying turbulence. Therefore, we can presume that the benefits of its unconventional reduced space are truly visible only when the PDE system is complex enough. In addition to the distance metric, more thorough analysis of latent space contents via, e.g., perceptual metrics, also remains our future work.

Moreover, our method has proven its capabilities in scenarios where force fields were inferred by our networks besides the velocity fields. In the forced turbulence case, the forces were external factors that were independent from the velocity data, thus our \emph{ATO} model had no difficulty finding a latent representation that led to a superior performance. In Sec.~\ref{sec:smoke_plume}, we showed that our model gave promising results in a scenario where the forces were internal, i.e. created by a marker field that was dependent of the latent velocity. That case opens interesting future work, such as finding the best reduced representations for the coupled marker and velocity fields.

Although we evaluated our model on various scenarios, its generalization for broader applications still remains a challenge. As our model allows for the efficient production of high-resolution simulations with a reduced solver, it is potentially attractive for editing physics simulations within the learned reduced space in real-time. Indeed, once a coarse initial frame is transformed into \emph{ATO}'s latent space, it is easy to tweak the physical properties of the reduced solver (e.g., viscosity) or to add external factors, such that it can produce high-resolution simulations in a more interactive way. Accordingly, the adaptation of our \emph{ATO} setup for three-dimensional problems is a promising topic for future work.

\section{Conclusion}
\label{sec:conclusion}

We have presented \emph{ATO}, a model that leverages interactions between neural networks and a differentiable physics solver to autonomously explore reduced representations for high-resolution fluid restoration purposes. Our results show that deep neural networks can learn to develop new dynamics for specific learning objectives by using the simulated degrees of freedom as latent space. Our approach opens the path to the exploration of physical latent spaces for other PDEs, as well as different learning tasks than the restoration of details of fluid simulations.

\section{Acknowledgments}
This project has been funded by the Futur \& Ruptures PhD program of the \emph{Fondation Mines-Telecom}.


\printbibliography                

\setcounter{section}{0}
\setcounter{figure}{0}
\setcounter{equation}{0}

\twocolumn[\centering \LARGE \textbf{Appendix}\\\vspace{3em}]

\section{Implementation details}

We provide the implementation details with the code and data in the supplemental material, along with the neural network architectures. We refer to the guideline (i.e., \emph{README.md}) to reproduce the results presented in this work. The code and trained models of our experiments will be published upon acceptance.

\section{Experiments}\label{appx:experiments}

In order to acquire a training data-set for each scenario, we generate a set of solution sequences of the given PDE problem.
The PDEs from our experiments work with a continuous velocity field $\mathbf{v}$ in the two-dimensional space, i.e., $\mathbf{v} = [v_x, v_y]^T$. Considering reference simulations on regularly discretized grids, we focus on exploring latent spaces (i.e., reduced representations) that are four times coarser than the reference.

\subsection{Karman vortex street}\label{appx:karman}

\begin{figure*}[hb]
  \centering
    \includegraphics[width=\textwidth]{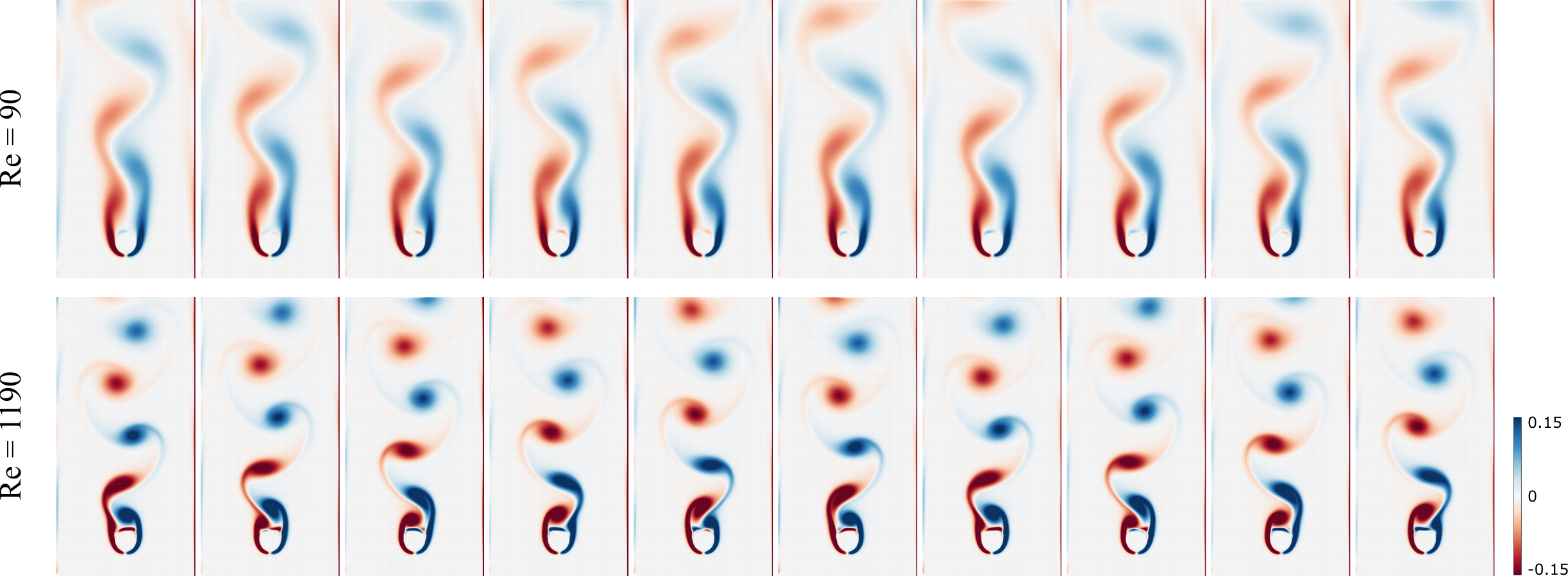}
    \caption{Two examples from the training data-set of the Karman vortex street scenario: Re = 90 and Re = 1190.}
 \label{appx:fig:karman_dataset}
\end{figure*}

This first example targets a complex PDE problem within a constrained setup, where the velocity field evolves over time while being constrained to be divergence free. We evaluate the incompressible Navier-Stokes equations for Newtonian fluids:
\begin{equation}
\label{appx:eq:ns}
\frac{\partial{\mathbf{v}}}{\partial{t}} = - (\mathbf{v} \cdot \nabla)\mathbf{v} -
\frac{\nabla{p}}{\rho} + \nu \nabla^2\mathbf{v}
\quad \textrm{subject to} \quad \nabla \cdot \mathbf{v} = 0
\end{equation}
where $p$ is the pressure, $\rho$ is the density, and $\nu$ is the viscosity
coefficient. The reference simulation domain is discretized with 128$\times$256
cells and a cell spacing of one using a staggered grid scheme. We use closed boundary conditions for the sides and open boundary conditions for the top of the domain; at the bottom, we set a constant inflow velocity. The continuous inflow collides with a fixed circular obstacle, which creates an unsteady wake flow that evolves differently depending on the Reynolds number. For the temporal discretization, the unit time step size is used. We generate 20 simulations of 200 steps each and randomly choose 5\% of them for the validation set and the remaining 95\% for the training set. We use Reynolds numbers in \{90, 120, 140, 150, 160, 170, 180, 190, 200, 220, 290, 340, 390, 490, 540, 590, 690, 740, 790, 1190\}, and we skip the first 2000 time-steps in order to let the flow stabilize. Both the least and most turbulent simulations of the training set are shown in Fig.~\ref{appx:fig:karman_dataset}.

In order to make our training more stable, we use pre-trained networks with eight integrated steps as warm starts for our final models. Each training uses 100 epochs with a batch size of ten. The learning rate starts from 4$\times$10$^{-4}$ and exponentially decays with a decaying rate of 0.9 every ten epochs. If divergence happens while training, we restart our training with a smaller learning rate. In this example, we compare all the models trained with 16 integrated steps. We also note that the encoder and adjustment models of \emph{ATO}, the corrector of \emph{SOL}, and the solver of \emph{Dil-ResNet} take the Reynolds number as additional input.

\begin{figure*}[htb]
  \centering
    \includegraphics[width=\textwidth]{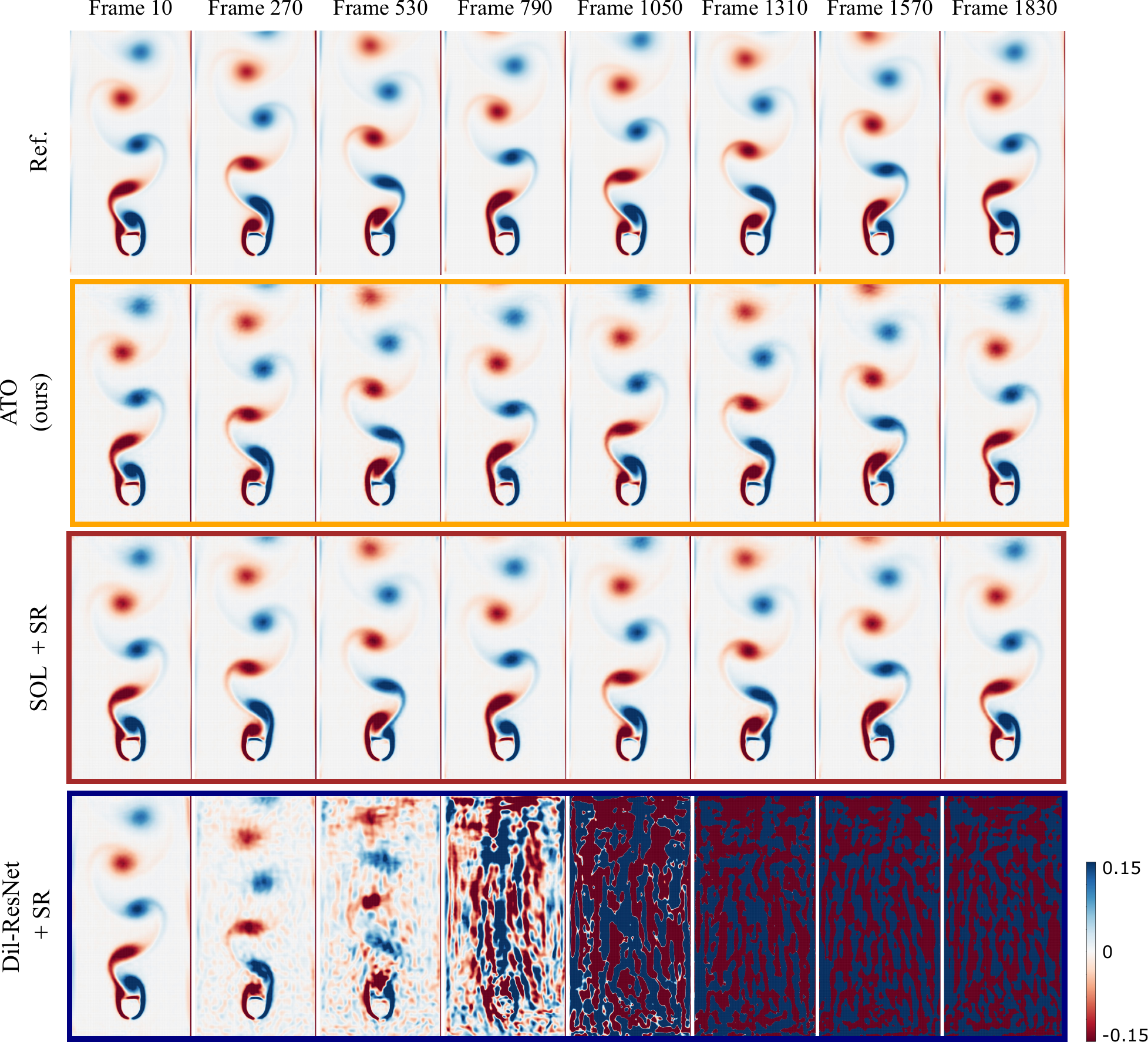}
    \caption{Restored frames of different models for the Karman vortex street scenario with Re = 850.}
 \label{appx:fig:karman_test}
\end{figure*}

\begin{figure*}[htb]
  \centering
    \includegraphics[width=\textwidth]{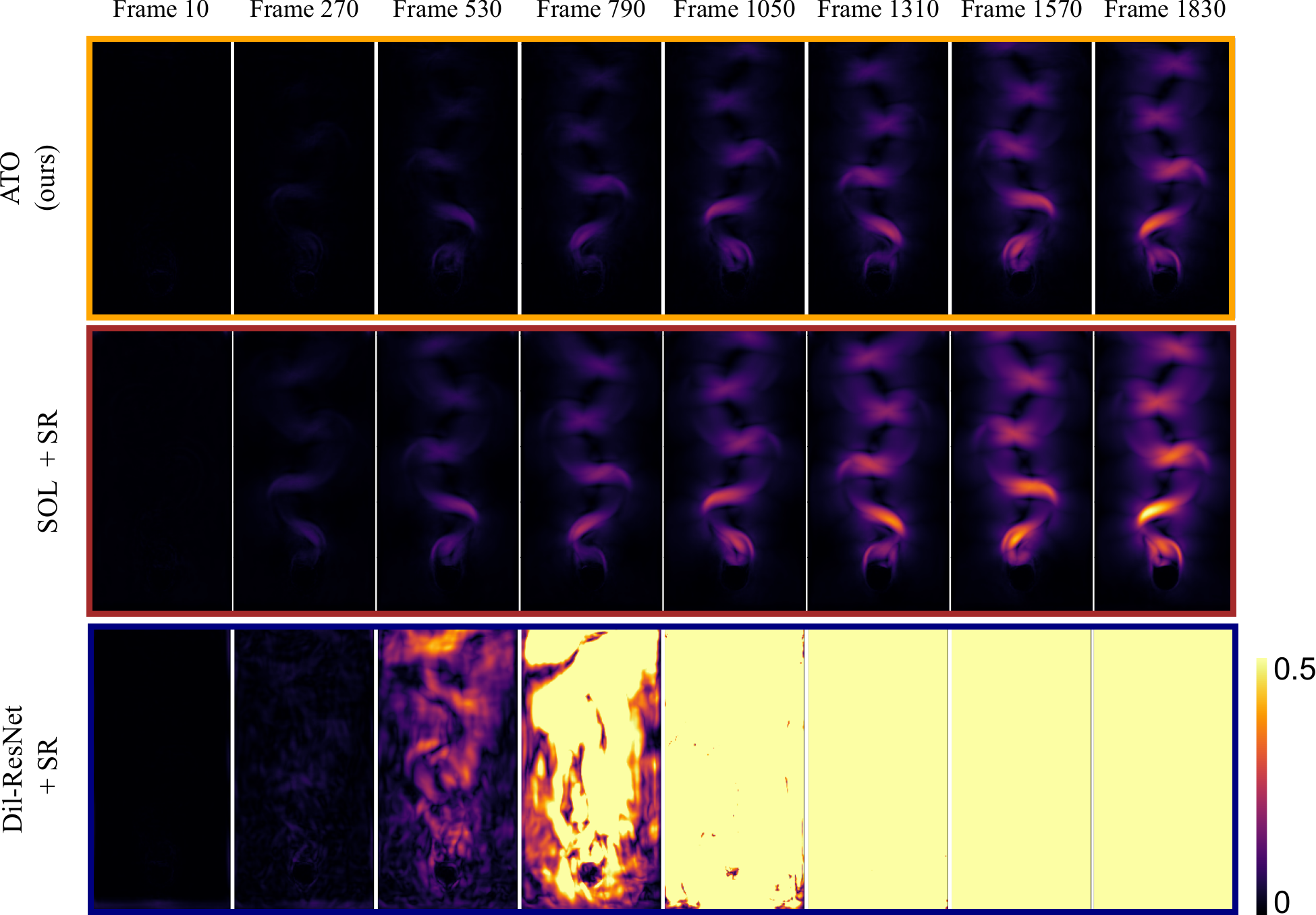}
    \caption{Absolute error in velocity for the different models, for the Karman vortex street scenario with Re = 850.}
 \label{appx:fig:karman_error_maps}
\end{figure*}

The test set consists of six solution trajectories evaluated with Reynolds numbers in \{450, 650, 850, 1050, 1200, 1400\}. Example sequences of the test data and the inference results of different models are shown in Fig.~\ref{appx:fig:karman_test}, for $Re=850$, along with the spatial distribution of the velocity error in Fig.~\ref{appx:fig:karman_error_maps}.

\begin{figure*}[htb]
  \centering
  \includegraphics[width=0.49\linewidth]{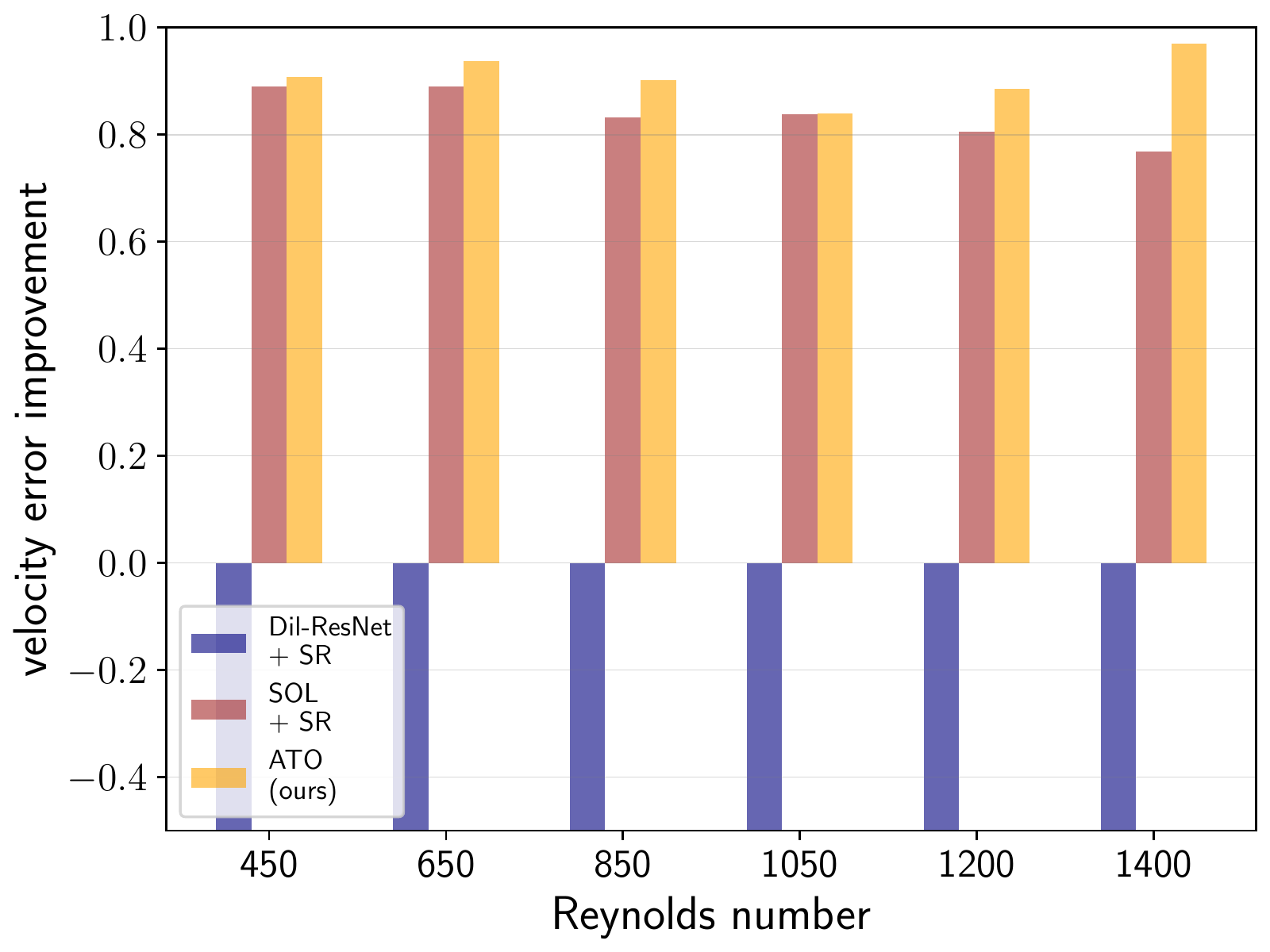}
  \includegraphics[width=0.49\linewidth]{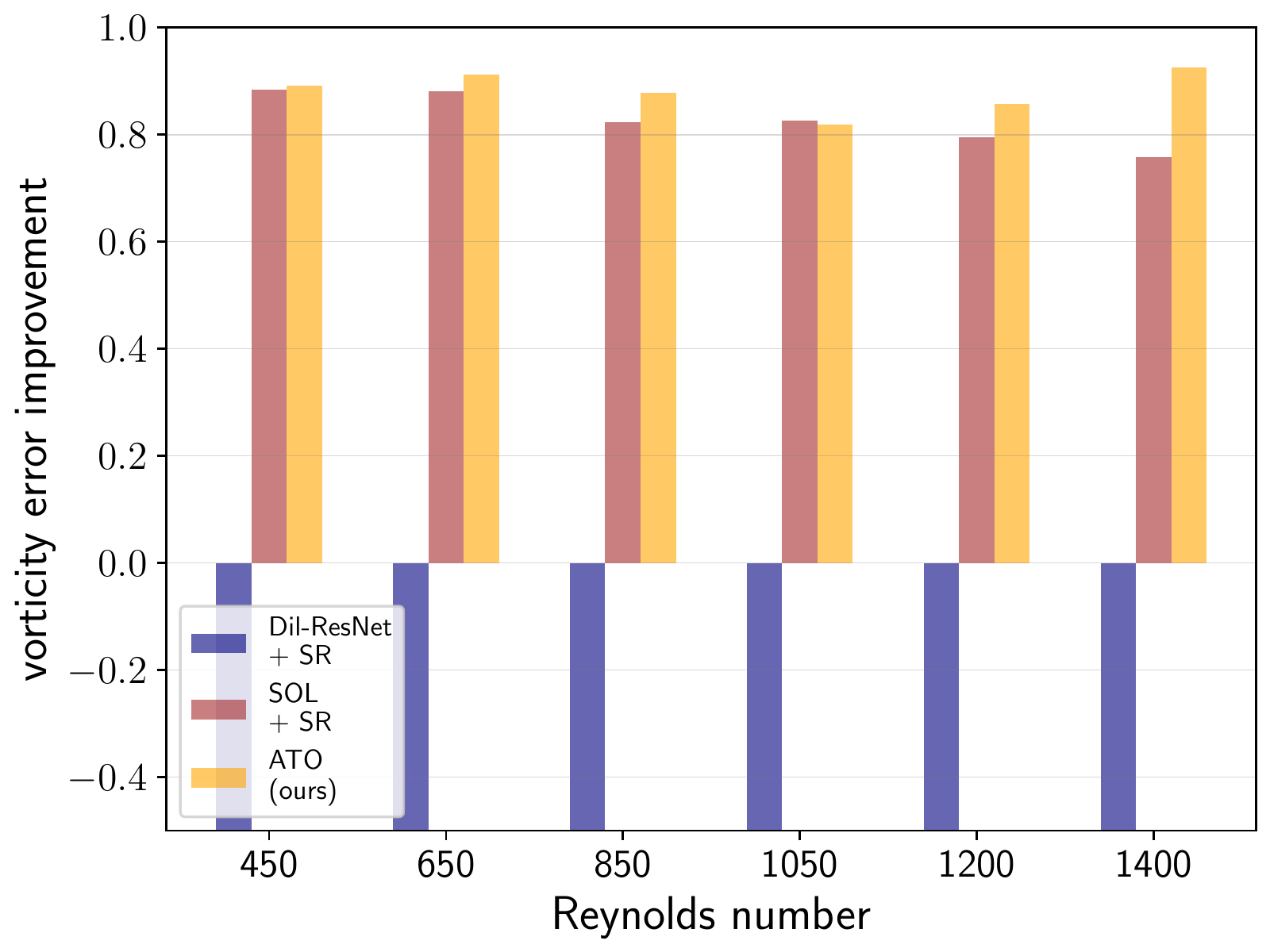}
  \caption{Velocity (left) and vorticity (right) error improvements for six different Reynolds numbers between 450 and 1400. The highest Reynolds number used for training is 1190. The \emph{ATO} model generalizes better than the others.}
  \label{appx:fig:karman_imp}
\end{figure*}

\begin{figure*}[htb]
  \centering
  \includegraphics[width=0.49\linewidth]{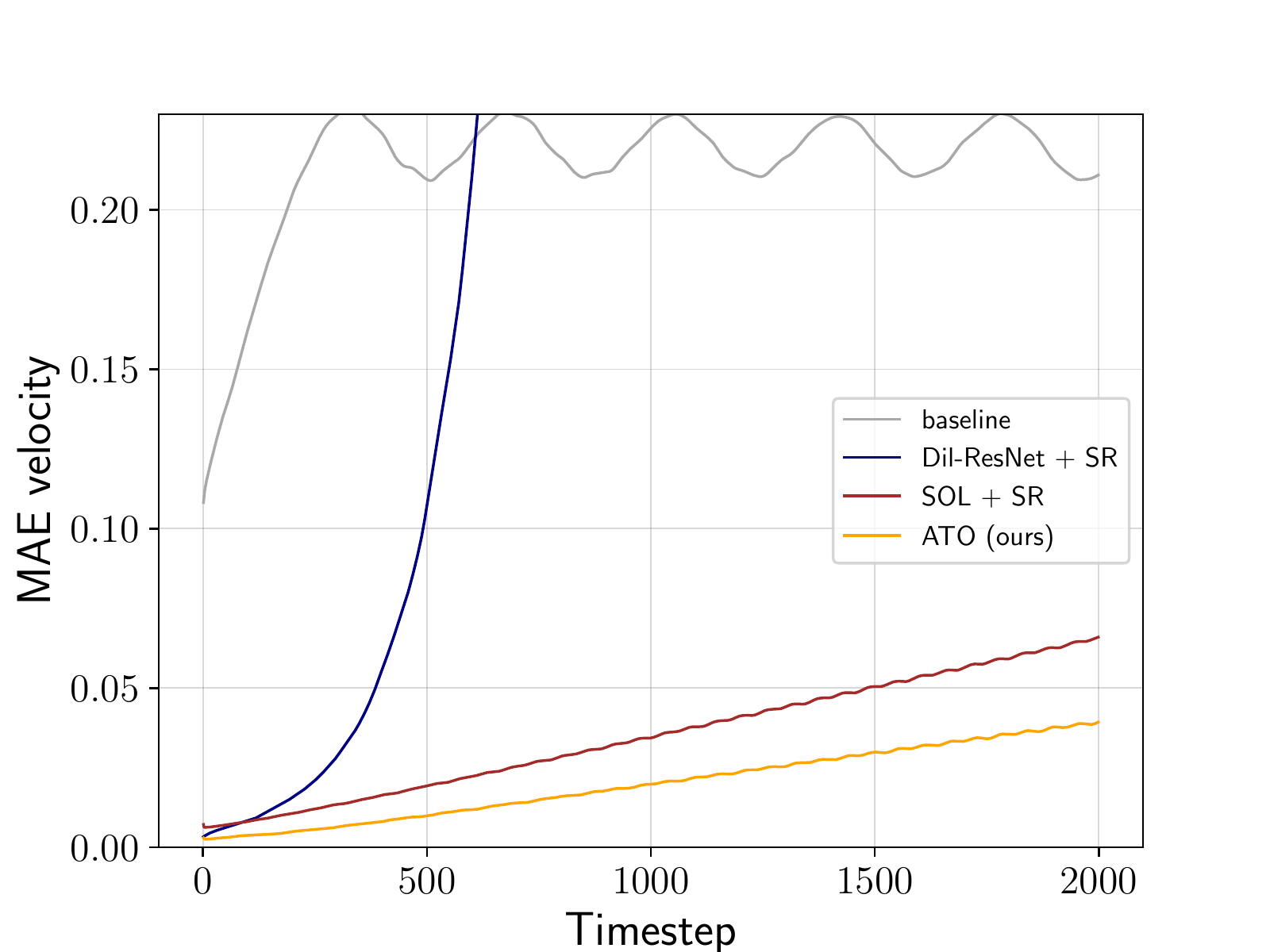}
  \includegraphics[width=0.49\linewidth]{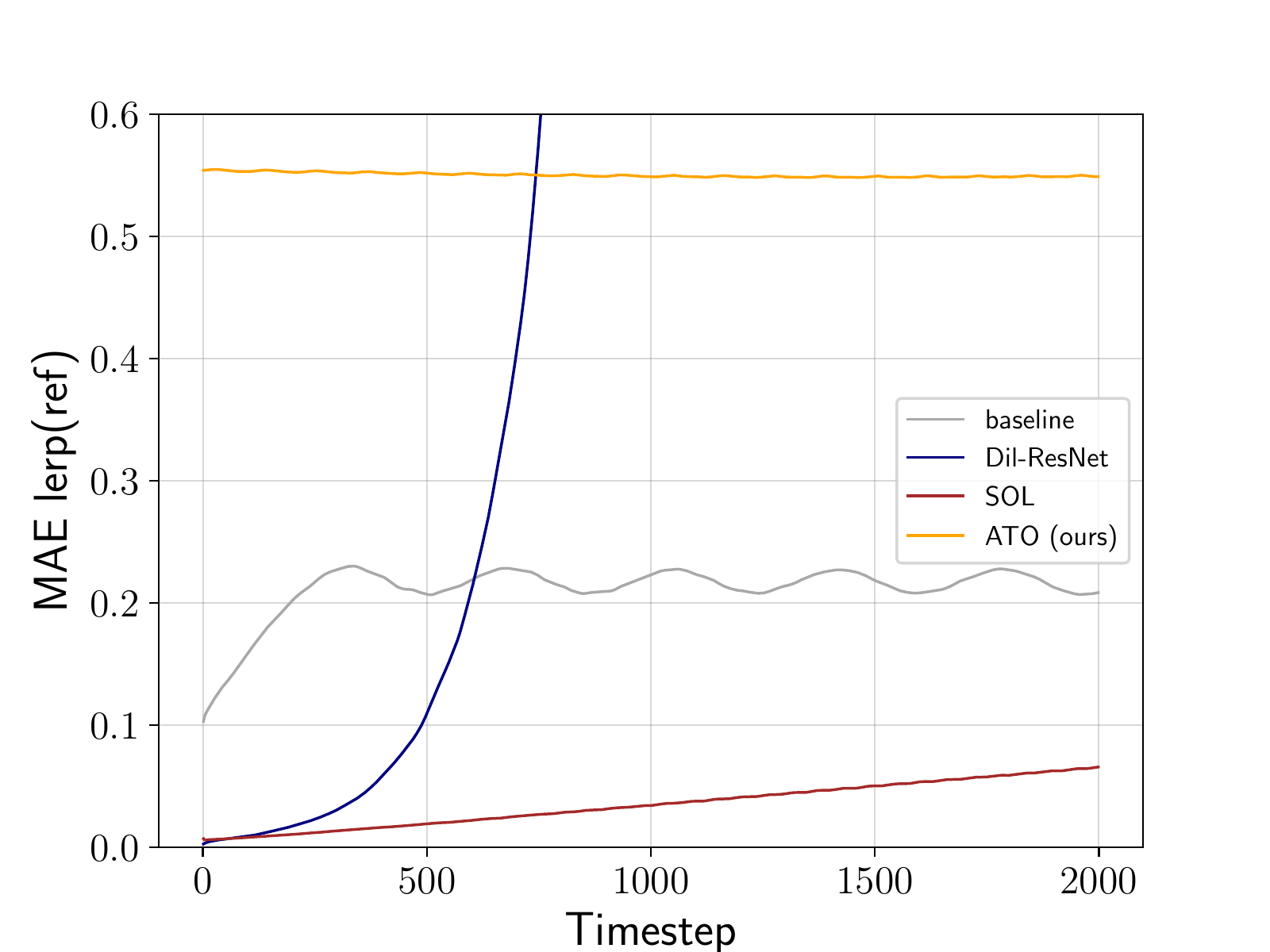}
  \caption{MAEs of recovered velocities (left) and distances of the reduced
    spaces to the down-sampled reference (right) over time for the Karman vortex street scenario.}
  \label{appx:fig:karman_time}
\end{figure*}

As can be seen on Fig.~\ref{appx:fig:karman_imp}, which shows the velocity and vorticity error improvements of each model over the baseline, \emph{ATO} presents the best generalization capabilities. Fig.~\ref{appx:fig:karman_time} shows the temporal evolution of the velocity MAE and the distance between each model's reduced space and \emph{lerp(ref)}.

\subsection{Decaying turbulence}\label{appx:decaying-turb}

\begin{figure*}[htb]
  \centering
  \includegraphics[width=\textwidth]{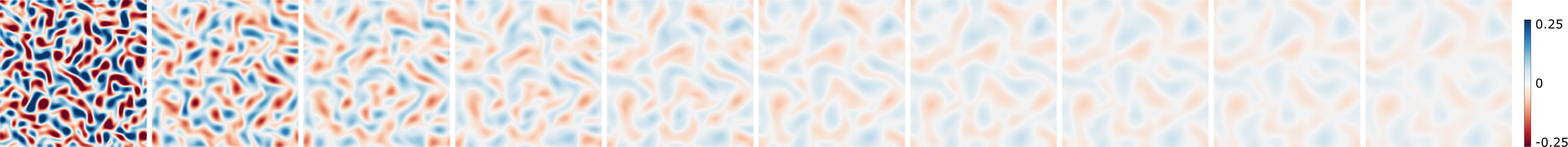}
  \caption{Example frames from one simulation of the training data-set of the decaying turbulence scenario.}
 \label{appx:fig:decaying-turb-dataset}
\end{figure*}

This example tackles the same incompressible Navier-Stokes equations, but with vortices intialized all over the physical space that slowly decay over time. In this scenario, the viscosity stays constant (equal to $0.1$) within the training and test data-sets. The reference simulation domain is discretized with 128$^2$ cells and a cell spacing of one. Both the discrete velocity and pressure values are stored at the center of each cell, and periodic boundary conditions are applied. For the temporal discretization, a time step size of 1.0 is used. The training data-set consists of 20 simulations of 200 steps each, which evolve from different initial velocity fields. We use randomly selected 5\% for the validation set and the remaining 95\% for the training set. An example sequence of the data is shown in Fig.~\ref{appx:fig:decaying-turb-dataset}. As in the Karman vortex street case, we first train our models with eight integrated steps as warm starts for the final models. We compare all the models trained with 16 integrated steps.

\begin{figure*}[htb]
  \centering
    \includegraphics[width=\textwidth]{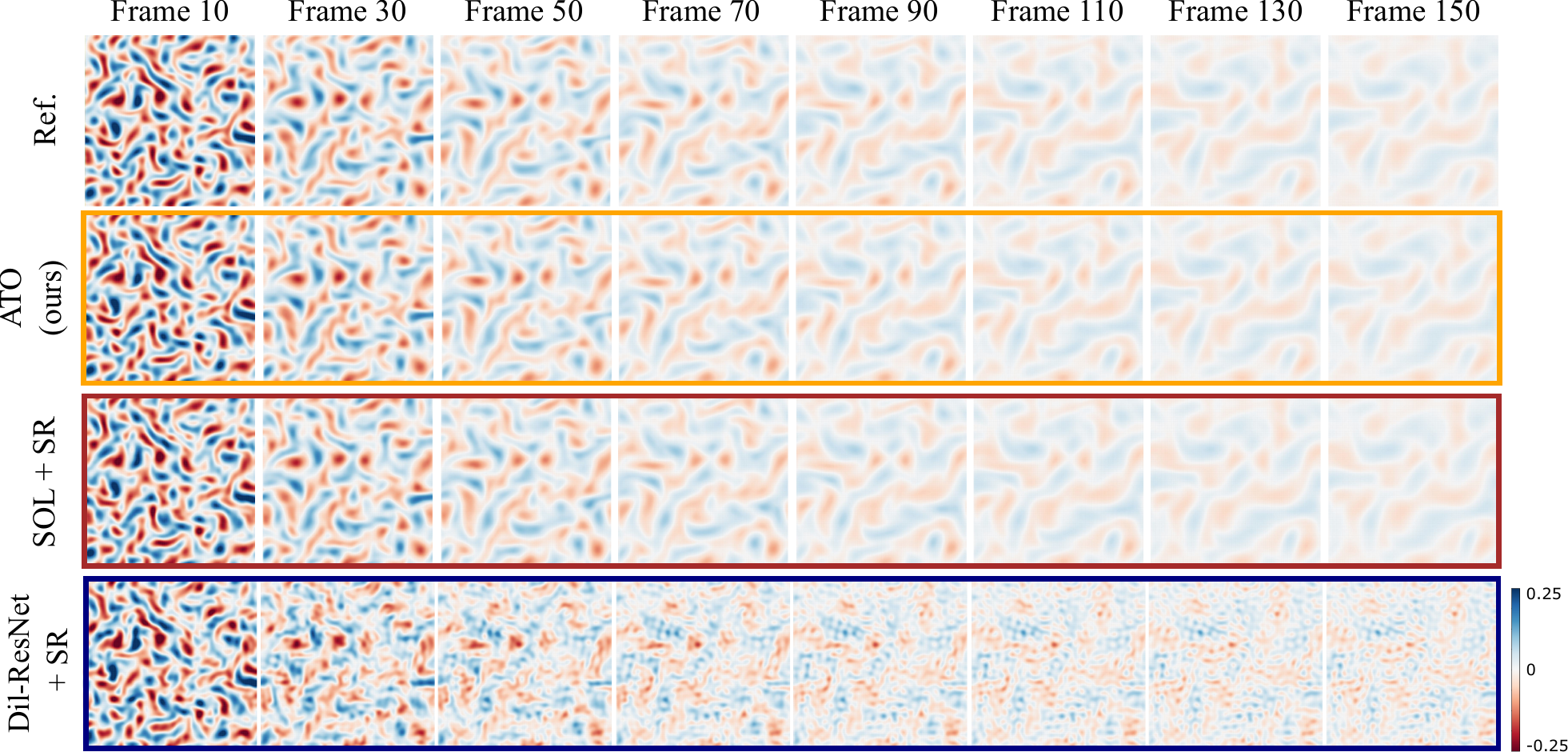}
    \caption{Example frames of a test case for different models for the decaying turbulence scenario.}
 \label{appx:fig:decaying-turb-test}
\end{figure*}

\begin{figure*}[htb]
  \centering
   \includegraphics[width=\textwidth]{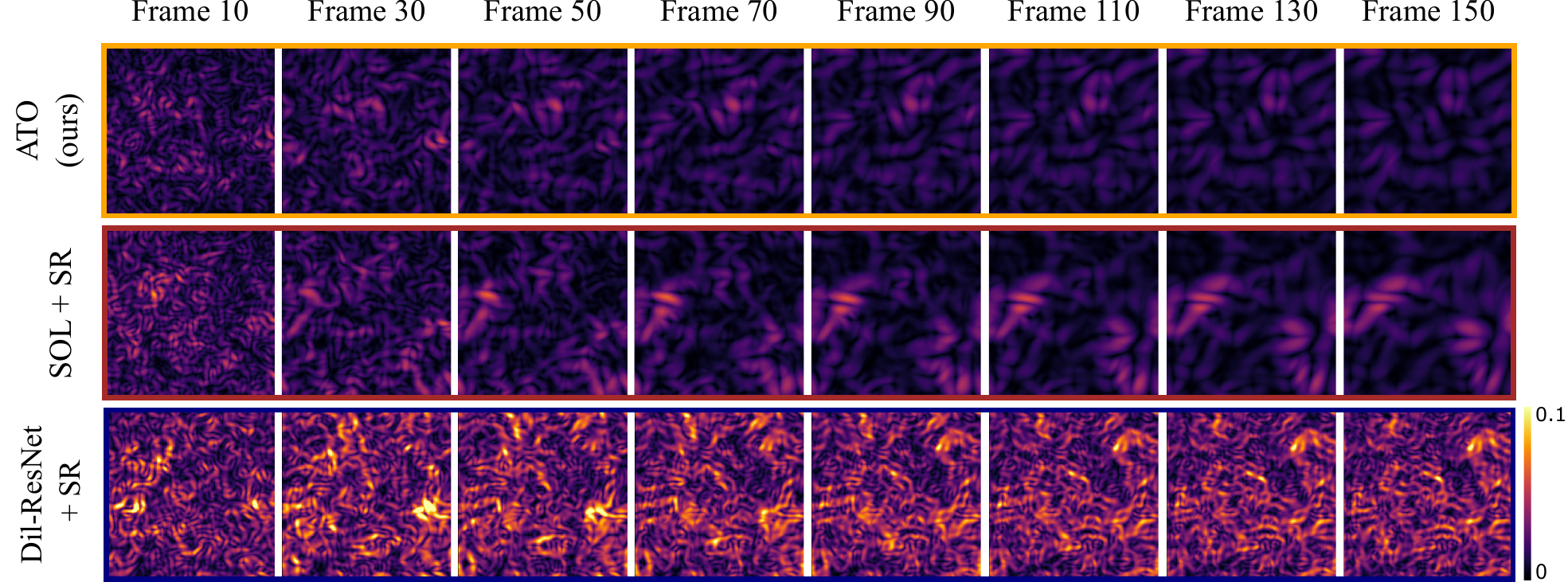}
  \caption{Absolute error in velocity for the different models, for the decaying turbulence scenario.}
  \label{appx:fig:decaying-turb-error-maps}
\end{figure*}

\begin{figure*}[htb]
  \centering
   \includegraphics[width=0.49\textwidth]{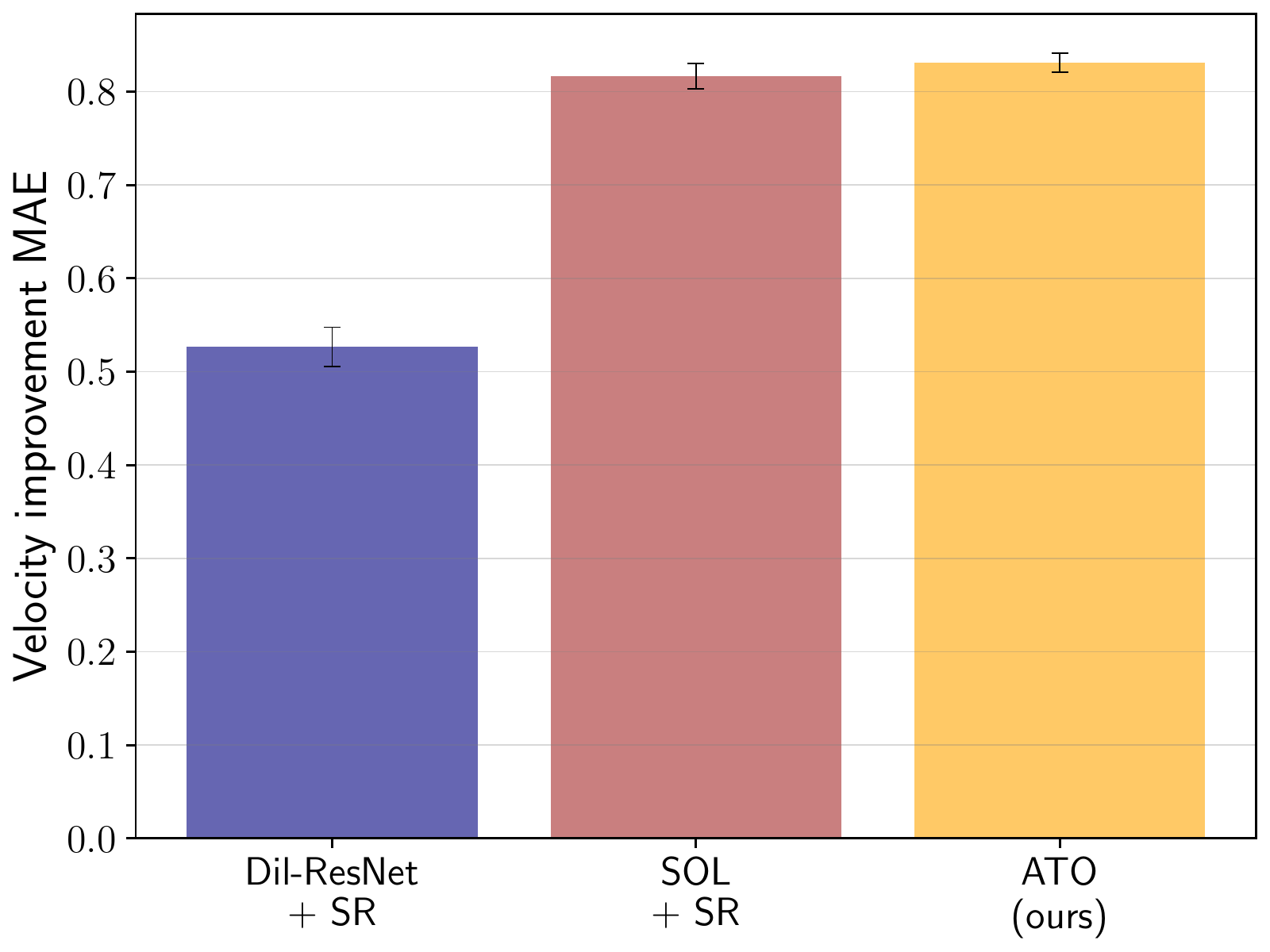}
   \includegraphics[width=0.49\textwidth]{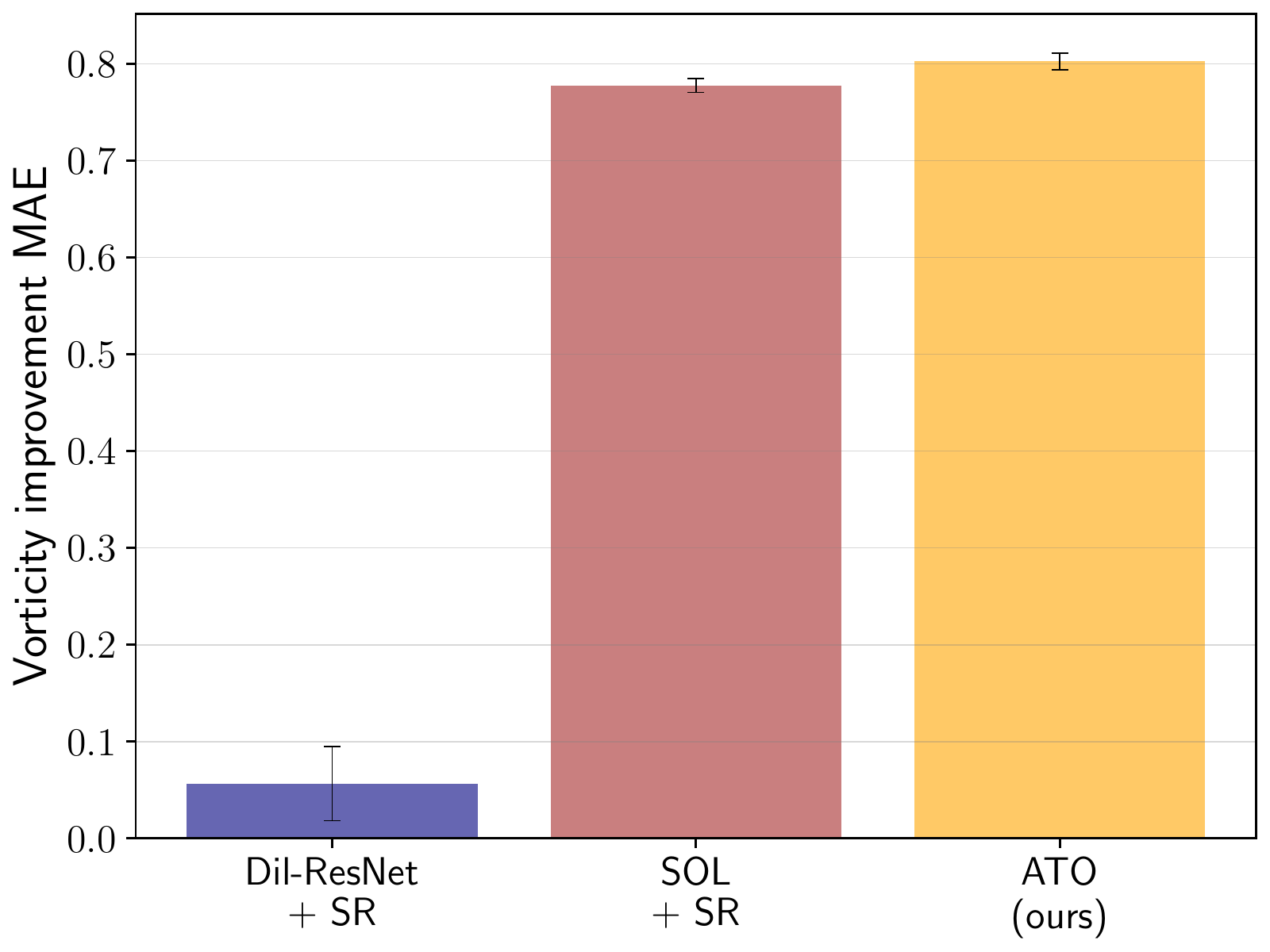}
  \caption{Velocity (left) and vorticity (right) error improvements for the decaying turbulence scenario. The \emph{ATO} model improves the baseline the most for every test case.}
  \label{appx:fig:decaying-turb-imp}
\end{figure*}

\begin{figure*}[htb]
  \centering
   \includegraphics[width=0.49\textwidth]{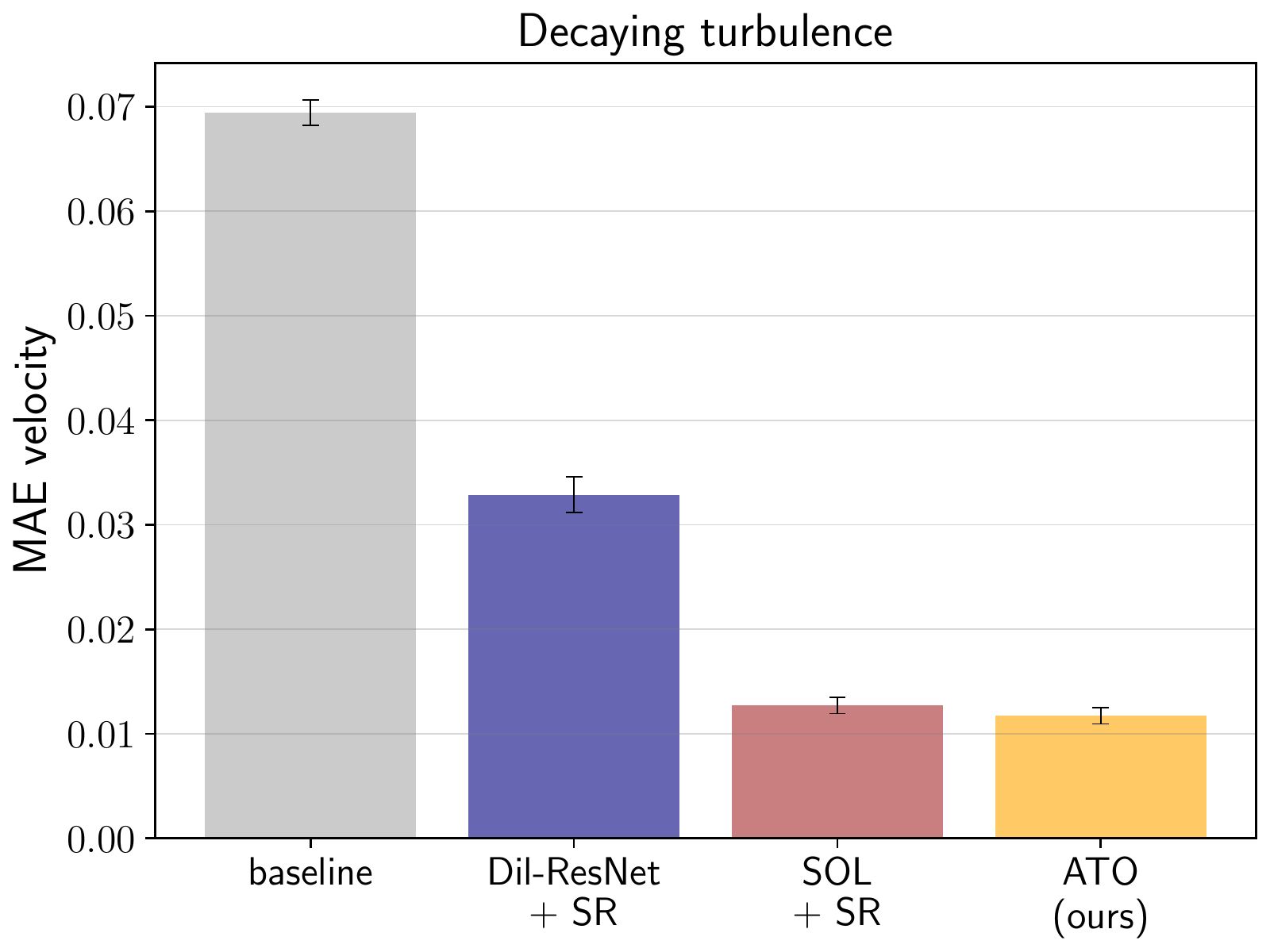}
   \includegraphics[width=0.49\textwidth]{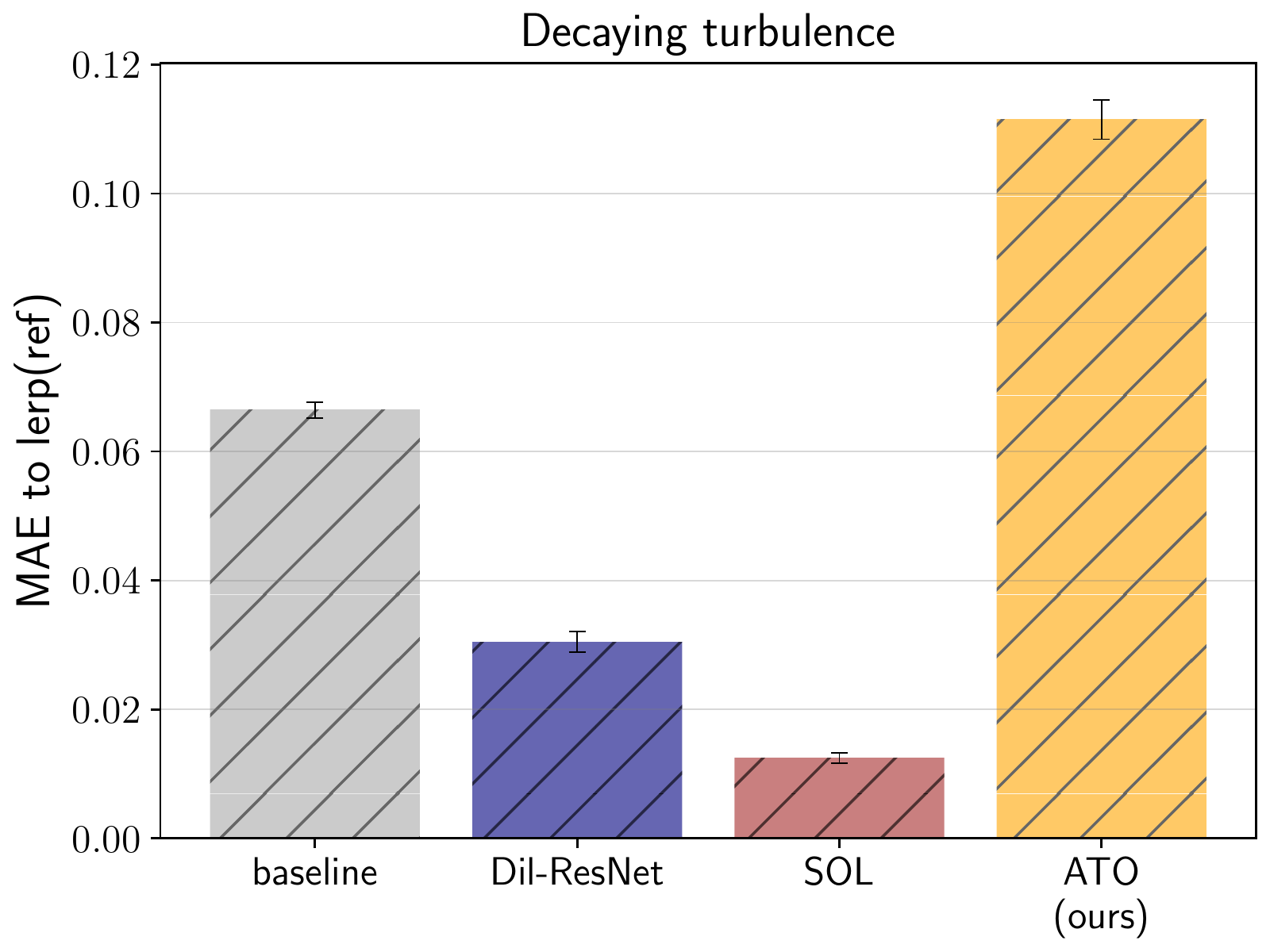}
  \caption{MAEs of recovered velocities (left) and distances of the reduced
    spaces to the down-sampled reference (right) for the decaying turbulence scenario.}
  \label{appx:fig:decaying-turb-mae}
\end{figure*}

\begin{figure*}[htb]
  \centering
 \includegraphics[width=0.49\linewidth]{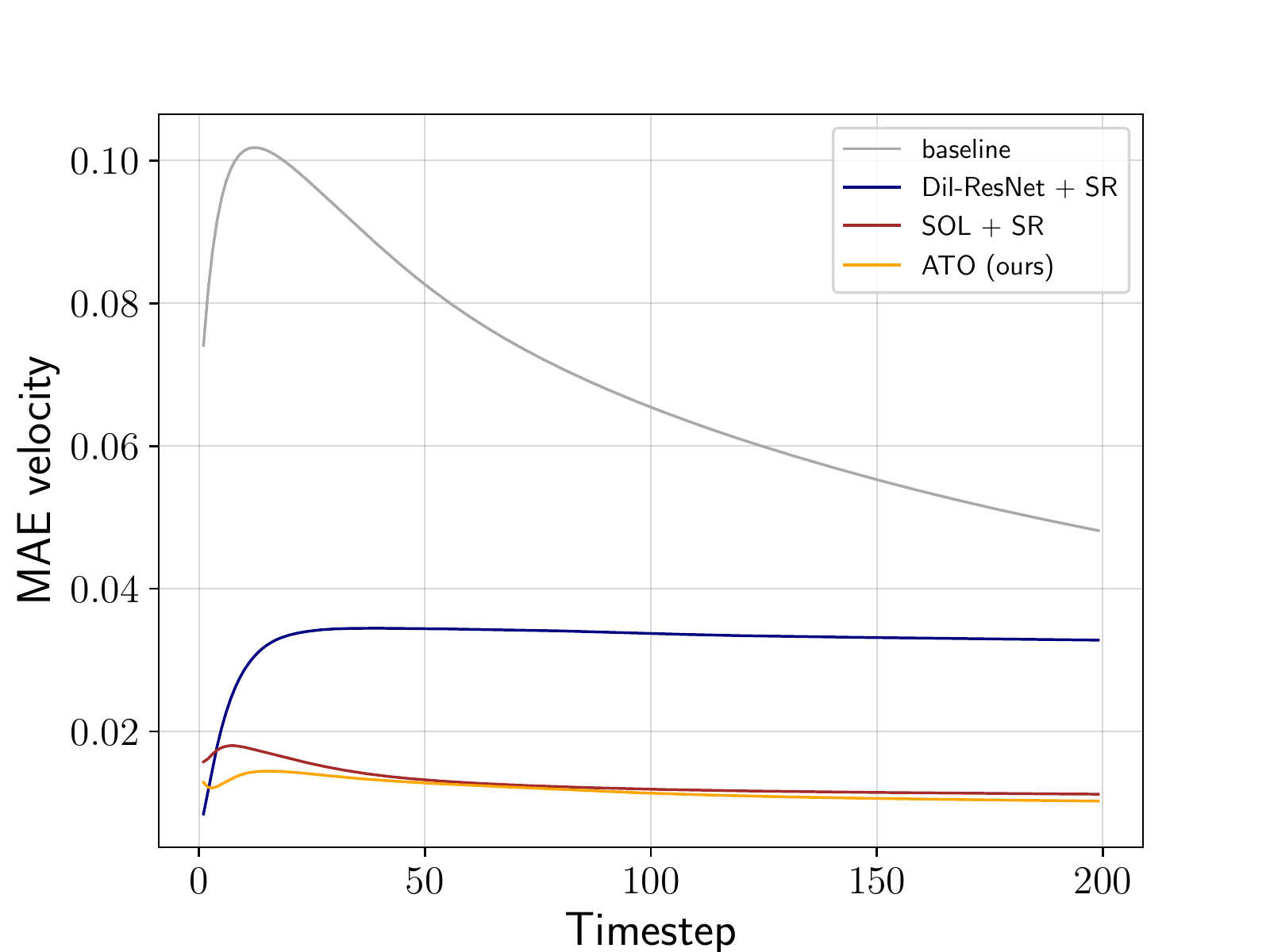}
  \includegraphics[width=0.49\linewidth]{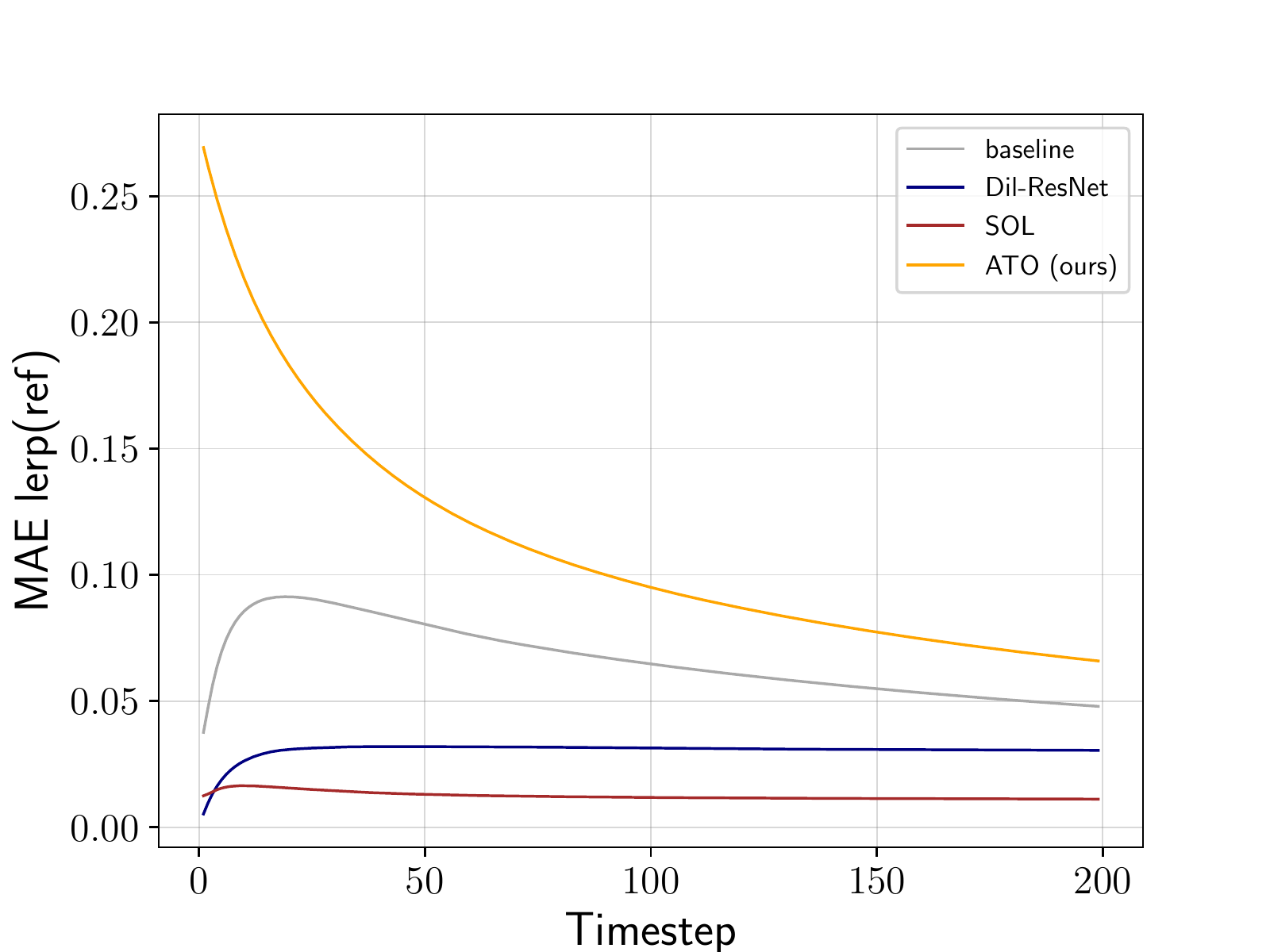}
  \caption{MAEs of recovered velocities (left) and distances of the reduced
    spaces to the down-sampled reference (right) over time for the decaying turbulence scenario.}
  \label{appx:fig:decaying-turb-time}
\end{figure*}

The five test trajectories evolve from different initial velocity fields on a domain identical to the training one. Fig.~\ref{appx:fig:decaying-turb-test} shows the inference results of the different models for one example, and Fig.~\ref{appx:fig:decaying-turb-error-maps} shows the spatial distribution of the error for the same example.
Fig.~\ref{appx:fig:decaying-turb-imp} and Fig.~\ref{appx:fig:decaying-turb-mae} (left) show that our \emph{ATO} model improves the baseline the most in both velocity and vorticity metrics, although \emph{SOL + SR} shows a comparable performance. As shown in Fig.~\ref{appx:fig:decaying-turb-time} and Fig.~\ref{appx:fig:decaying-turb-mae} (right), its latent space representation is more distant from the linearly down-sampled representation than the other models', yet it shows a similar or better performance.

\subsection{Forced turbulence}\label{appx:forced-turb}

This case has the same experimental setup as the previous one, but with an external force sequence $\mathbf{g}(\mathbf{x}, t)$ that is added to Eq.~(\ref{appx:eq:ns}). This force sequence yields complex, chaotic evolutions of vortices over time. We use a different force sequence for each simulation trajectory, composed of 20 overlapping sine functions as follows:
\begin{equation}
\begin{aligned}
\label{appx:eq:extforce}
  g_x(\mathbf{x}, t) &= \sum_{i=1}^{20}a_i\sin(k_i\alpha_i \cdot \mathbf{x} + w_i t + \phi_i) \\
  g_y(\mathbf{x}, t) &= \sum_{i=1}^{20}a_i\sin(k_i\alpha_i \cdot \mathbf{x} + w_i t + \phi_i)
\end{aligned}
\end{equation}
where $a_i$ is the amplitude, $k_i$ is the wave number, $\alpha_i$ is the wave
direction, $w_i$ is the frequency, and $\phi_i$ is the phase shift.
These values are randomly sampled from uniform distributions as follows:
$a_i \in [-0.1, 0.1]$, $k_i \in \{6, 8, 10, 12\}$, $w_i \in [-0.2, 0.2]$, and
$\phi_i \in [0, \pi]$. $\alpha_i$ is a random angle ($\in [0, 2\pi]$).
The composed sine functions are, then, evaluated over the domain mapped into $[0, 2\pi]$ for each dimension.

\begin{figure*}[htb]
  \centering
  \includegraphics[width=\textwidth]{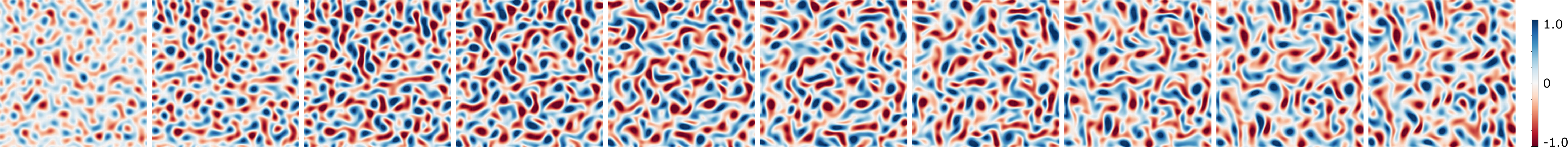}
  \caption{Example frames from one simulation of the training data-set of the forced turbulence scenario.}
 \label{appx:fig:forced-turb-dataset}
\end{figure*}

For the temporal discretization, a time step size of 0.2 is used. The training data-set consists of 20 simulations of 200 steps each, which evolve from different initial velocity fields with different force sequences. We use randomly selected 5\% for the validation set and the remaining 95\% for the training set. An example sequence of the data is shown in Fig.~\ref{appx:fig:forced-turb-dataset}. We use the models trained on the previous decaying turbulence case as warm starts for our final models, trained for 100 epochs. We compare all the models trained with 16 integrated steps. The encoder of \emph{ATO} shares its weights for velocity and force in order to learn a unified operation for both reduced representations.

\begin{figure*}[htb]
  \centering
    \includegraphics[width=\textwidth]{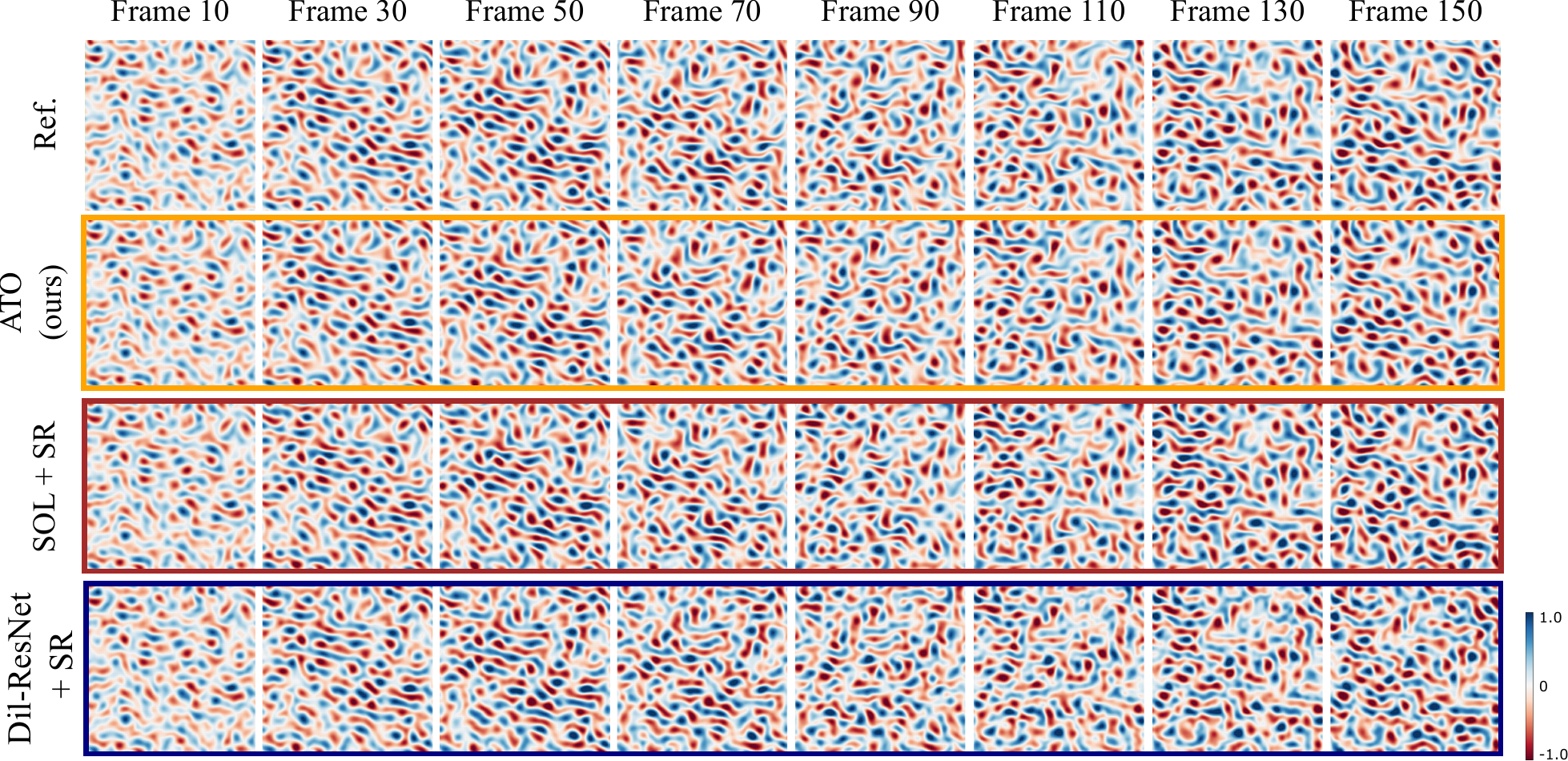}
    \caption{Example frames of a test case for different models for the forced turbulence scenario.}
 \label{appx:fig:forced-turb-test}
\end{figure*}

\begin{figure*}[htb]
  \centering
   \includegraphics[width=\textwidth]{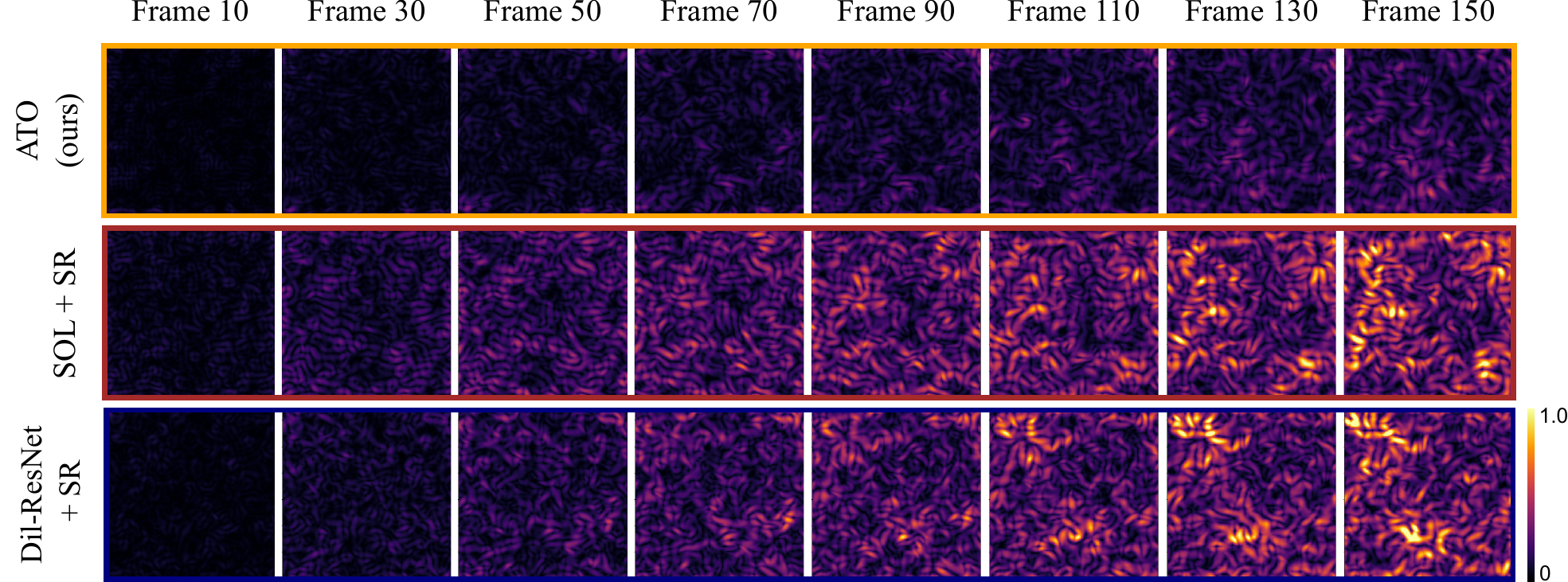}
  \caption{Absolute error in velocity for the different models, for the forced turbulence scenario.}
  \label{appx:fig:turb-error-maps}
\end{figure*}

\begin{figure*}[htb]
  \centering
   \includegraphics[width=0.49\textwidth]{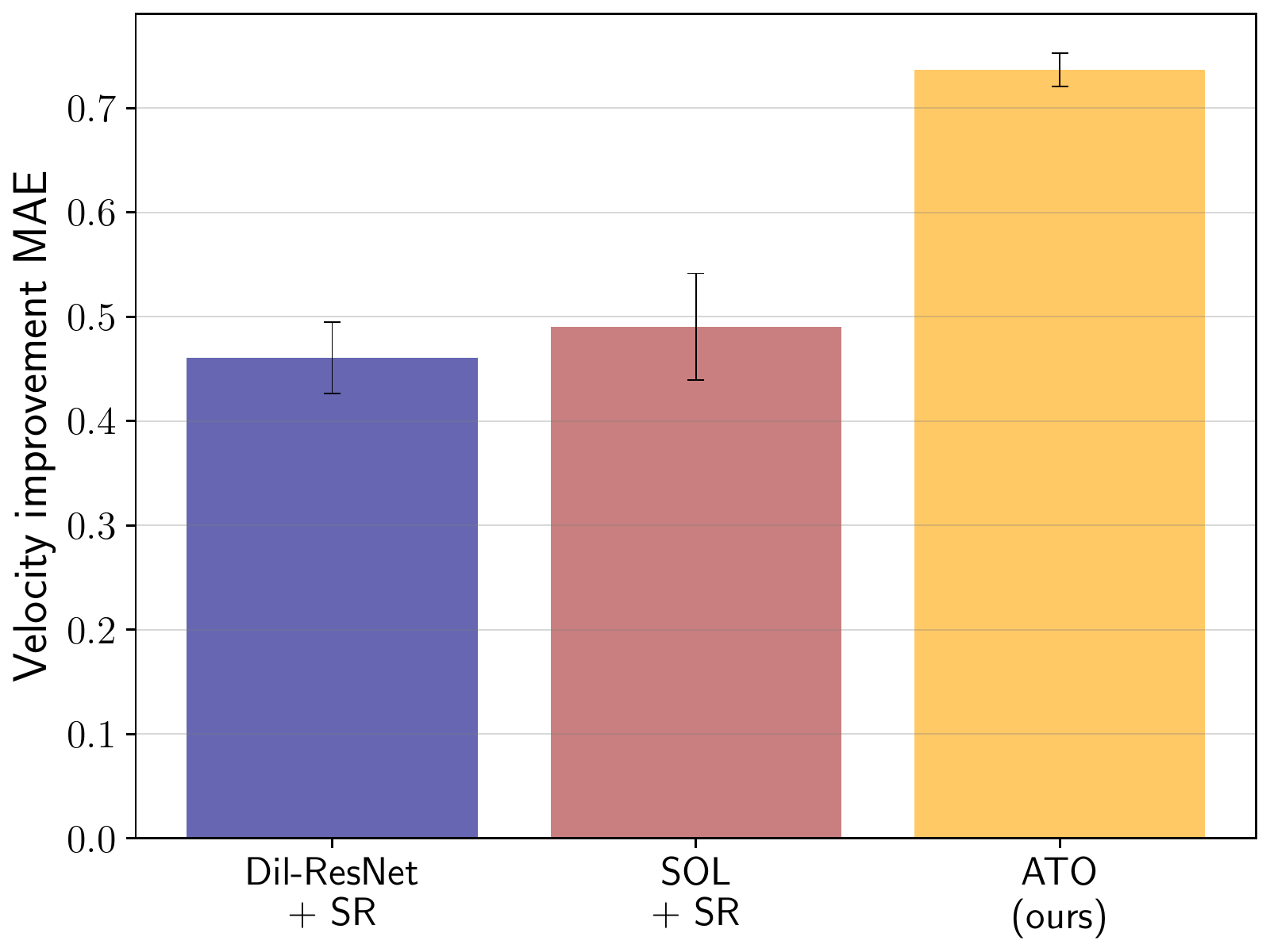}
   \includegraphics[width=0.49\textwidth]{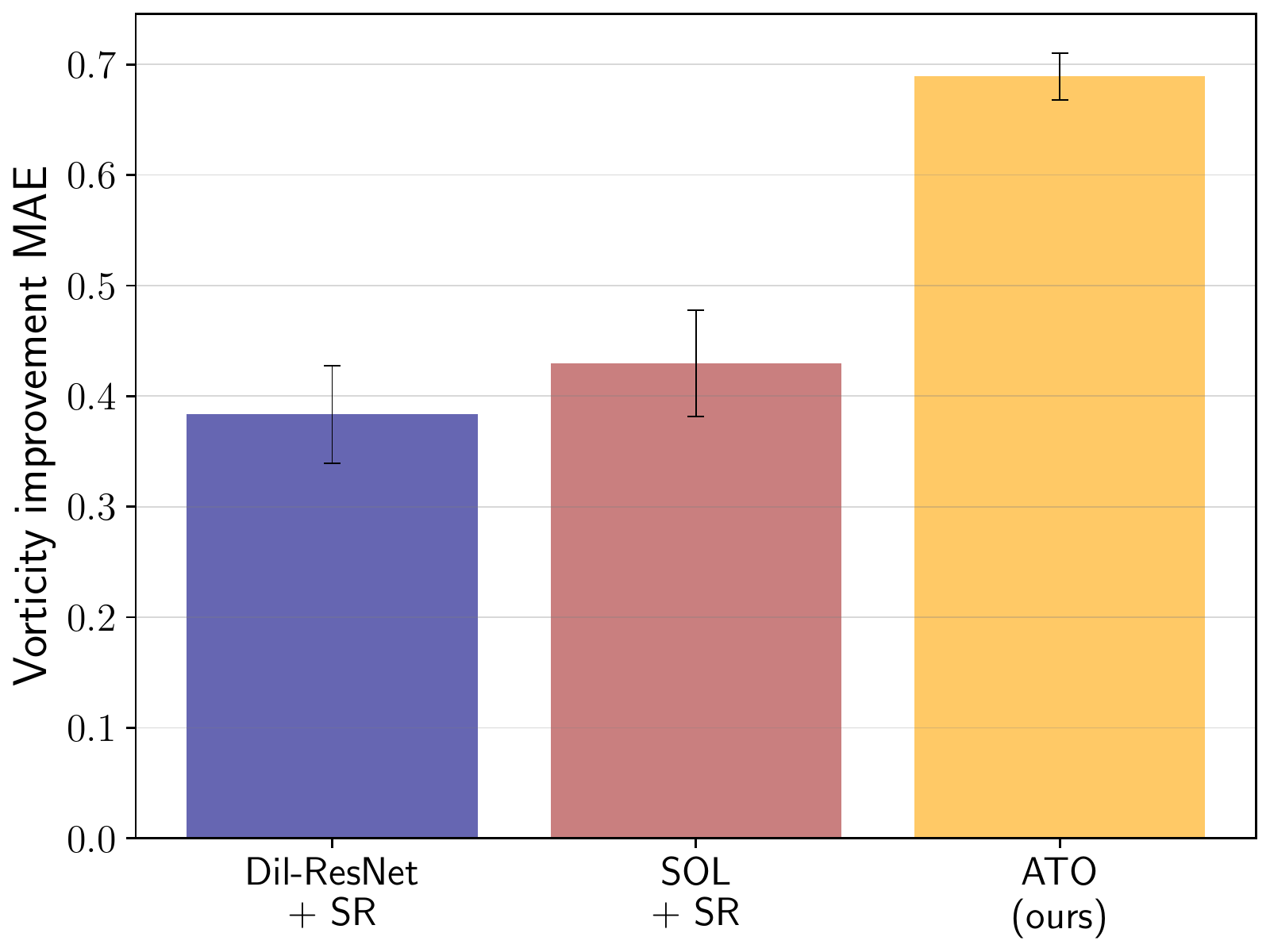}
  \caption{Velocity (left) and vorticity (right) error improvements for the forced turbulence scenario. The \emph{ATO} model improves the baseline the most for every test case.}
  \label{appx:fig:forced-turb-imp}
\end{figure*}

\begin{figure*}[htb]
  \centering
  \includegraphics[width=0.49\linewidth]{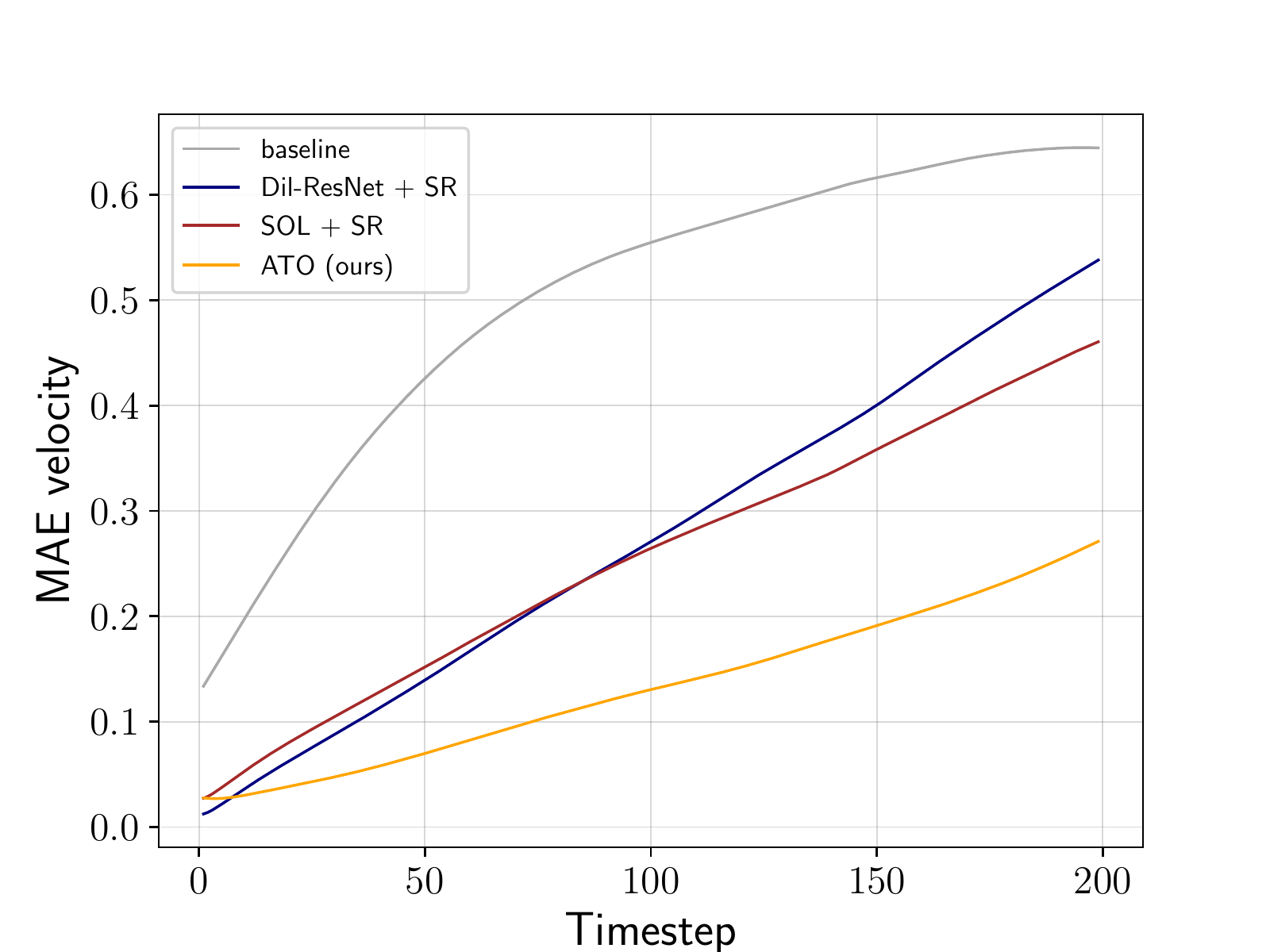}
  \includegraphics[width=0.49\linewidth]{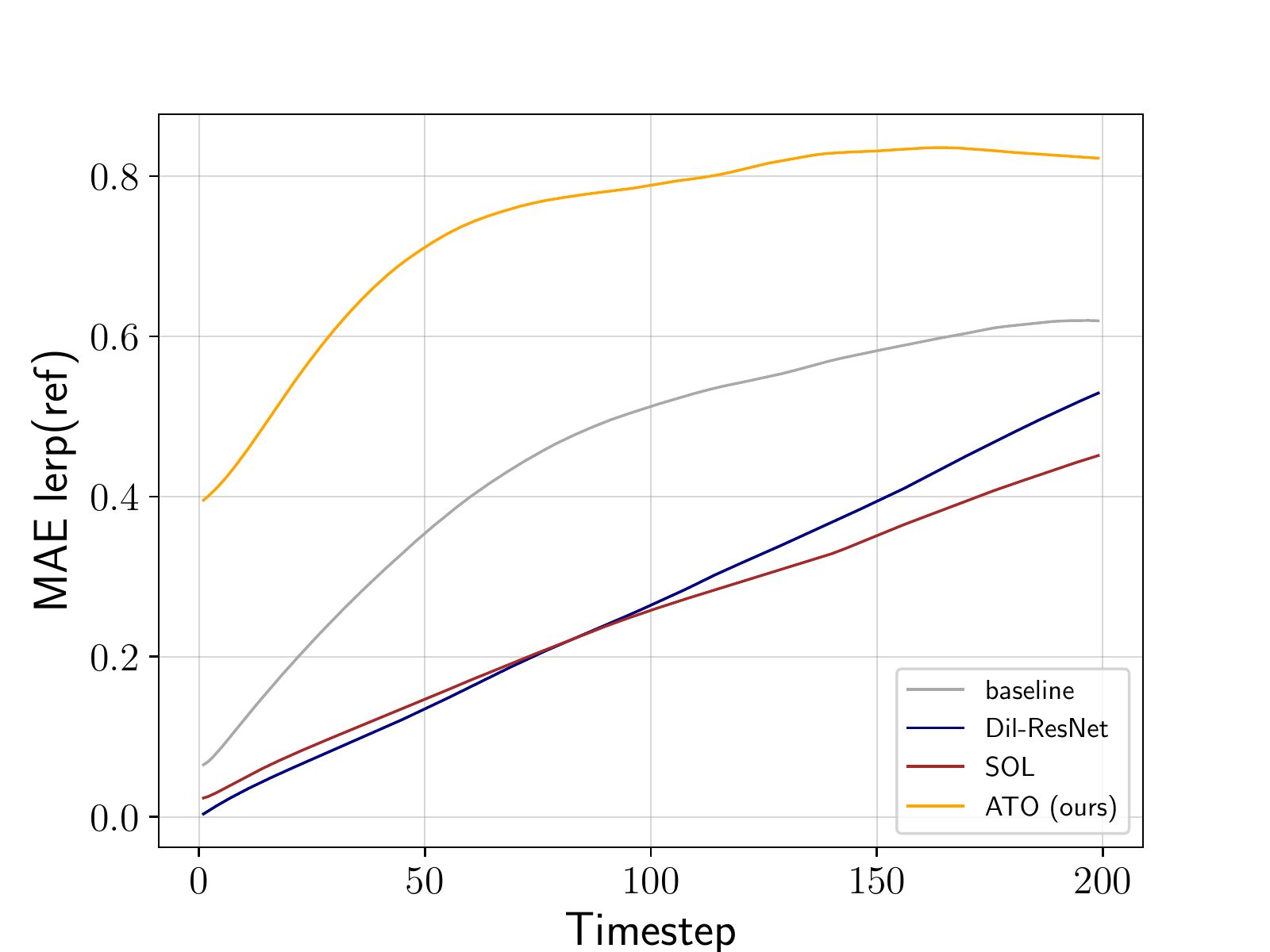}
  \caption{MAEs of recovered velocities (left) and distances of the reduced
    spaces to the down-sampled reference (right) over time for the forced turbulence scenario.}
  \label{appx:fig:forced-turb-time}
\end{figure*}

The five test trajectories evolve from both different initial velocity fields and different force field sequences.
Fig.~\ref{appx:fig:forced-turb-test} shows the inference results of the different models for one example, and Fig.~\ref{appx:fig:turb-error-maps} shows the spatial distribution of the error for the same example.
Fig.~\ref{appx:fig:forced-turb-imp} shows that our \emph{ATO} model improves the baseline the most in all five test cases in both velocity and vorticity metrics, while its latent space representation (Fig.~\ref{appx:fig:forced-turb-time}) is more distant from the linearly down-sampled representation than the other models'.

\subsection{Smoke plume}\label{appx:smoke}

\begin{figure*}[htb]
  \centering
  \includegraphics[width=\textwidth]{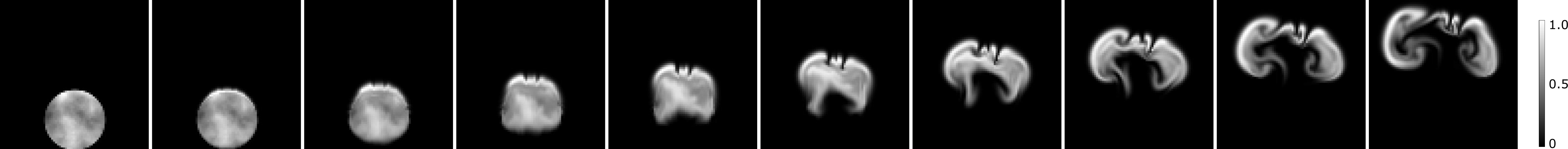}
  \caption{Example frames from one simulation of the training data-set of the smoke plume scenario.}
 \label{appx:fig:smoke-plume-dataset}
\end{figure*}

\begin{figure*}[htb]
  \centering
    \includegraphics[width=\textwidth]{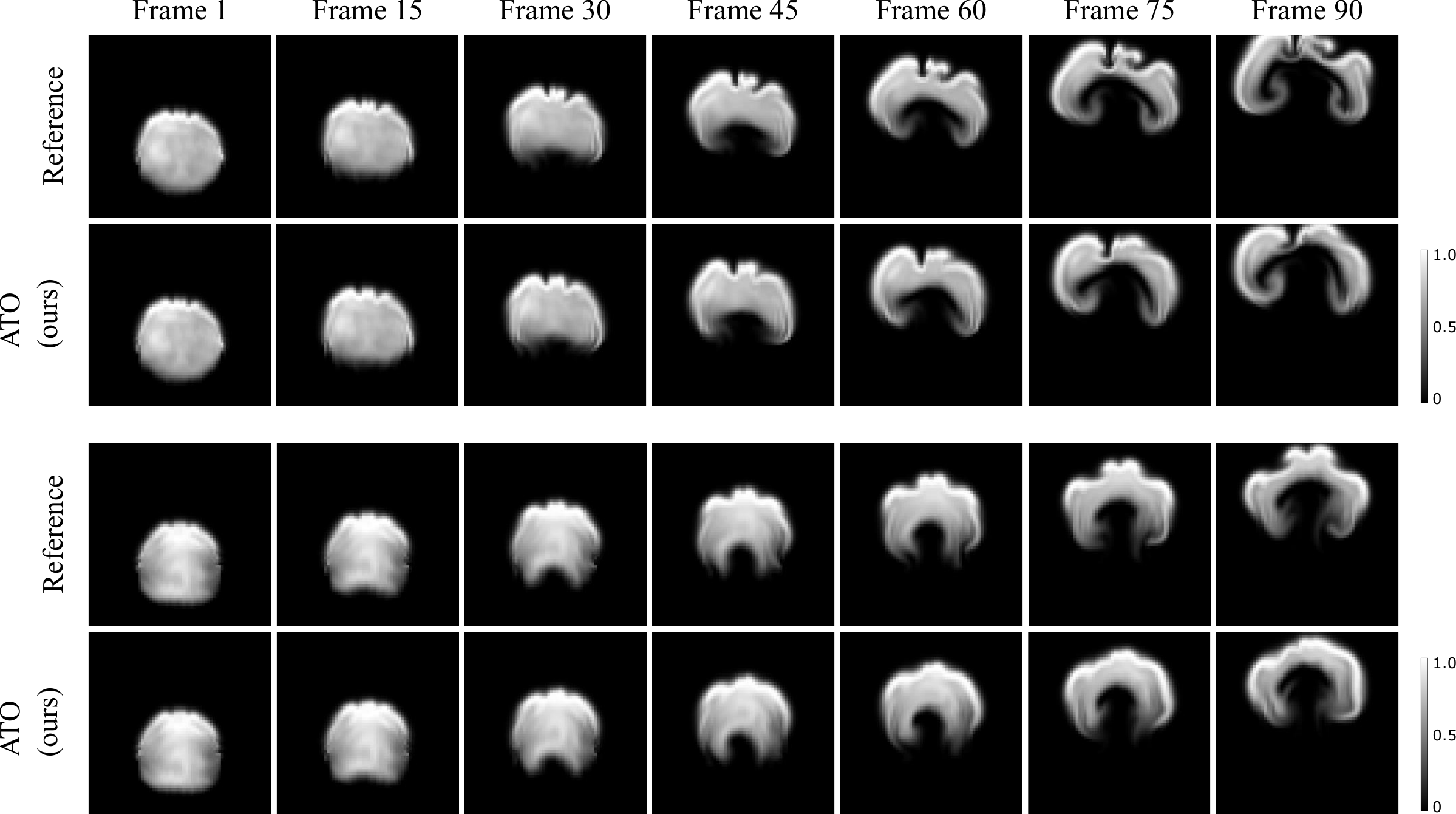}
    \caption{Example frames from our \emph{ATO} model for two test cases of the smoke plume scenario.}
 \label{appx:fig:smoke-plume-test}
\end{figure*}

This example represents a smoke volume of a circular shape, which is slowly rising up producing interesting swirling motions. A buoyancy force is produced by a passive marker field with a buoyancy factor of $0.25$ applied vertically. The training data-set consists of 20 simulations of 200 steps each with a time-step of $0.2$, which start from circular marker fields with a constant radius of $0.12$, but evolve differently due to the random initialization of the markers.
We use randomly selected 5\% for the validation set and the remaining 95\% for the training set. An example sequence of the data is shown in Fig.~\ref{appx:fig:smoke-plume-dataset}. For this case, we use more integrated solver steps than the others. We first train our model with four, eight, and 16 integrated steps as warm starts for the final model. We apply our \emph{ATO} model trained with 32 steps, for 100 epochs. Fig.~\ref{appx:fig:smoke-plume-test} shows the inferences of our \emph{ATO} model for different initializations.

\section{Neural Network Architectures}\label{appx:nn-arch}

In this section, we detail our network architectures for each model.
We note that the practical implementations of all the models can be found in the
supplemental code.

The encoder of the \emph{ATO} setup consists of two convolutional layers with 32 and 16 features each with a kernel size of five. Each convolutional layer is followed by the Leaky ReLU activation function. A last layer with two features and the same kernel size but without the activation infers the final encoded output. This model has approximately 15k trainable weights.

The adjustment of the \emph{ATO} setup and the corrector of \emph{SOL} (\cite{um2020solverintheloop}) employ an identical network model. This model consists of a first convolutional layer with 32 features and a kernel size of five, followed by five blocks of two convolutional layers with 32 features each and a kernel size of five. Each layer is followed by the Leaky ReLU activation function, and each block is connected to the next with a skip-connection. A last layer with two features follows with the same kernel size yet without the activation. This architecture has approximately 260k trainable weights.

The decoder used for the \emph{ATO} setup and the super-resolution model from \emph{Dil-ResNet + SR} and \emph{SOL + SR} are adapted from the multi-scale architecture of \cite{fukami2019super}, such that the total number of trainable weights is close to 97k.

The \emph{Dil-ResNet} model is adapted from the architecture of the state-of-the-art network model proposed for turbulent flow problems \cite{stachenfeld2022learned}. This model has a first convolutional layer with 32 features with a kernel size of three and no activation. It is followed by four identical blocks of seven convolutional layers with 32 features each with a kernel size of three and varying dilation rates from one to eight (respectively: 1, 2, 4, 8, 4, 2, 1). Each layer is followed by the ReLU activation function, and each block is linked to the next via a skip-connection. A last layer with two features follows with the same kernel size yet without the activation. This model has a similar number of trainable weights as the adjustment model's (i.e., 260k). We note that a larger model did not improve the performance. Contrary to the other models, \emph{Dil-ResNet} is trained for only one step at a time and uses a \emph{MSE loss}.

For the Karman vortex street and smoke plume cases, we adopt zero-padding, and for the forced and decaying turbulence cases, which use periodic boundary conditions, we use periodic padding for all models.

\section{Training hyper-parameters}\label{appx:hyperparam}

In this section, we detail the choice of hyper-parameters for the different trained models.

Firstly, the code of the \emph{SOL} model from \cite{um2020solverintheloop} was released publicly, which enabled an easy reproduction of their experiments.
Secondly, the learning setup of the \emph{Dil-ResNet} model from \cite{stachenfeld2022learned} was precisely described in the article. We extensively tested the various hyper-parameters and chose the \emph{Dil-ResNet} model over \emph{Con-Dil-ResNet} (i.e., without the additional loss constraint) because it performed better in our physical scenarios. For training, we chose a Gaussian noise with $\sigma = 0.01$ for all scenarios.
Lastly, for our \emph{ATO} setup, we chose the depth of the models by making a compromise between performance and runtime/resources.
For the learning rate and batch size, since the physics solver made the models harder to train, we chose the values that best stabilized our training. Our loss being divided in two terms of different orders of magnitude (a high-resolution term and a low-resolution one), we set $\lambda_{hires}$ to $1$ for all scenarios and $\lambda_{latent} = 1$ for the Karman vortex
street case, $\lambda_{latent} = 1$ = 10 for the decaying turbulence, $\lambda_{latent} = 100$ for the forced turbulence and $\lambda_{latent} = 10$ for the smoke plume.

\end{document}